\documentclass[a4paper]{article}

\usepackage[english]{babel}
\usepackage[utf8x]{inputenc}
\usepackage[T1]{fontenc}

\usepackage[a4paper,top=3cm,bottom=2cm,left=3cm,right=3cm,marginparwidth=1.75cm]{geometry}

\usepackage{comment}
\usepackage{amsmath}
\usepackage{amssymb}
\usepackage{algorithm}
\usepackage{algorithmic}
\usepackage{multirow}
\usepackage{graphicx}
\usepackage{dcolumn}
\usepackage{bm}
\usepackage{float}
\usepackage[noblocks]{authblk}

\begin{document}
\title{Greedy Step Averaging: A parameter-free stochastic optimization method}
\author[*]{Xiatian Zhang}
\author[*]{Fan Yao}
\author[*]{Yongjun Tian}
\affil[*]{TalkingData Technology(Beijing)Co,.Ltd, China, \authorcr Email: \{xiatian.zhang, fan.yao, yongjun.tian\}@tendcloud.com}

\maketitle

\begin{abstract}
In this paper we present the greedy step averaging(GSA) method, a parameter-free stochastic optimization algorithm for a variety of machine learning problems. As a gradient-based optimization method, GSA makes use of the information from the minimizer of a single sample's loss function, and takes average strategy to calculate reasonable learning rate sequence. While most existing gradient-based algorithms introduce an increasing number of hyper parameters or try to make a trade-off between computational cost and convergence rate, GSA avoids the manual tuning of learning rate and brings in no more hyper parameters or extra cost. We perform exhaustive numerical experiments for logistic and softmax regression to compare our method with the other state of the art ones on 16 datasets. Results show that GSA is robust on various scenarios.
\begin{description}
\item[Keywords]
Optimization, algorithm, learning rate, parameter-free, self-adaptive, averaging strategy
\end{description}
\end{abstract}

\section{Introduction}

A large number of problems in machine learning involve minimizing the sum of a loss function over different training samples. The most prevalent algorithm for these problems is stochastic gradient descent method(SGD) \cite{SGD1, SGD2}. During a typical iteration step of SGD, the iterator randomly chooses a training instance, takes the negative gradient direction $-\frac{\partial l_i}{\partial \omega}$ of the local loss function $l_i$ as the descent direction, and moves forward with a specified step length, or learning rate $\lambda_{t}$. In addition to SGD, there is the standard full gradient(FG) method. Instead of using the gradient of the loss function of a single training sample, it employs evaluation of the total loss function. Although FG method enjoys a much better convergence performance, its formidable computational cost makes it intractable in large scale context: in each step we have to visit all training sample to obtain the full gradient. In industry communities, large datasets are usually under distributed storage and traverse over the whole dataset induce a considerable or even unaffordable input-output expenditure. As a matter of fact, in most industrial cases we only need to obtain a model with tolerable accuracy within limited time and memory. Thus, the SGD method becomes increasingly popular as an expedient alternative to FG method. 

Compared with FG, SGD enjoys greater efficiency in computation and implementation. However, SGD has an evident weakness: the tuning of learning rate $\eta$ is usually a heavy price to pay. Theoretically, to guarantee convergence we need to set learning rate $\eta_t$ as a decreasing sequence towards zero. However, the decreasing sequence is not well-grounded: why should we regard the training sample being scanned later less important than those being scanned earlier? In fact, since samples with brand new labels could appear at later stage of training process, we should not treat them with a comparatively small learning rate. And not to mention we have to specify the initial value as well as the decreasing rate of $\eta_t$. If we choose a constant step-size-strategy, the dilemma arises in another sense especially on large datasets: large $\eta$ leads to faster convergence but may inevitably bring in more fluctuations and a small $\eta$, while it may guarantee a smooth decreasing loss, often entails an unacceptably low convergence rate. In fact, the convergence rate and fluctuation of loss function, equivalent to the estimation bias and variance of parameter $\omega$, are intrinsically contradictive \cite{SGDTheo3} and we have to strike a balance between the two. Many attempts to address the dilemma including mini-batch strategy and many other variance reduction tactics have been made. In recent years, there have also emerged many self-adaptive methods like Adagrad, Adadelta, Adam. The ideas behind all these methods are to collect gradient information in previous iterations to calculate a preferable learning rate for the current step. However, all these methods need to specify an initial learning rate which is also a sensible parameter. Moreover, these methods are capable of dynamically adapting the learning rate by introducing new parameters. We have to tune these extra parameters to fit various circumstances. These methods can outperform traditional SGD in some scenarios. For example, Adadelta is claimed to be pretty effective for back propagation(BP) algorithm in neural network training.

However, there is another way to improve SGD: laying aside the learning rate selection issue and trying to alter the scheme to reach an accelerated linear convergence rate. There are already some breakthroughs, such as SAG, SVRG, SCSG algorithms. By exploiting history information of gradient and iteration sequence, these methods achieve a better performance. However, they introduce other sensitive parameters(SVRG, SCSG) and require intermittent full gradient evaluation and enormous memory occupation (SAG). And great efforts are needed to tune relevant parameters to guarantee convergence. Therefore, these attempts to improve SGD have encountered difficulties in most industrial cases.
  
For the reasons we have discussed above, we propose to enhance SGD in a new way which can maintain its greatest advantage of swiftness and at the same time avoid parameter tuning once and for all by introducing a practical step-size selection rule. This new method named as greedy step averaging(GSA) is a self-adaptive parameter free stochastic optimization algorithm for many machine learning problems, and works well for logistic and softmax regression in our experiment. It is equipped with a dynamically selected learning rate and maintains the same order of computational and memory cost.

The paper is organized as follows. In the next section, we will present some related works including most state of the art stochastic optimization algorithms which we will later compare with GSA. In the third and fourth section we introduce the heuristic idea and formulation of GSA. In the fifth section, we establish convergence analysis of the method. Finally, in section five and six, we conduct a comparison between GSA and the other methods mentioned above and offer some further discussions.
\section{Related Work}
There is a large variety of optimization methods derived from SGD utilizing merely the stochastic gradient information. We will not discuss algorithms that are infeasible in computation for high-dimensional data sets, e.g. second-order methods such as Newton’s method. Some of these methods claim to accelerate the convergence of SGD like Momentum, SAG, SVRG, at least in some specific circumstances while others claim using former iterative information to compute a optimal update. A full review of this immense literature would be outside the scope of our work. We only comment on the state of the art algorithms and several of the most widely-known ideas.

First we introduce the basic set up of SGD. Suppose our objective loss function is 
\begin{equation}
\text{minimize}_{\omega \in \mathbb{R}^p} \quad g(\omega) = \sum_{i=1}^{n}  f_{i}(\omega),
\end{equation}
where $\omega \in \mathbb{R}^p$ denotes the model parameter, $ f_{i}(\omega) = Loss(\omega, x_i, y_i)$ denotes the loss function with respect to $i$-th training sample $(x_i, y_i)$. 

The update scheme of $\omega$ is thus given by
\begin{equation}
\omega^{(k+1)} = \omega^{(k)} -\eta_{k} \nabla f_{i_k}(\omega^{(k)}).
\end{equation}

Denote $\omega^*$ the unique minimizer of $g(\omega)$. It has been proved that the full gradient method achieves a linear convergence rate:
$$
g(\omega^k)-g(\omega^*) = O(\rho^k),
$$
where $\rho$ is a constant depending on the condition number of $g$\cite{FGTheo}. 

The convergence performance of SGD is sublinear. Under standard assumptions, for SGD with constant step-size, one can obtain the unique minimizer with a linear convergence rate in a certain tolerance $C$:
$$
\mathbb{E}[g(\omega^k)] - g(\omega^*) = O(\rho^k) + C,
$$
where the expectation is taken with respect to all possible selection sequence of training samples, and $C$ is a constant depending on the condition number of $g$ and growing quadratically with $n$. To achieve a stable convergence behaviour, people often apply a diminishing step-size rule. For SGD under this strategy, we can reach a sublinear convergence rate\cite{SGDTheo1, SGDTheo2}:
$$
\mathbb{E}[g(\omega^k)] - g(\omega^*) = O(1/k).
$$

$\bm{Momentum}$

The Momentum method adds momentum term to SGD, the update scheme has the following form
$$
\omega^{(k+1)}=\omega^{(k)} - \eta_k \nabla f_{i_k}(\omega^{(k)})+ \beta_k(\omega^{(k)} - \omega^{(k-1)}).
$$
Momentum helps accelerate SGD in the relevant direction and dampens oscillations by adding a fraction $beta_k$ of the update vector of the past time step to the current update vector. The name Momentum demonstrates its property vividly: when move toward the approximated negative gradient direction in the $k+1-th$ step by $\eta_k$, the iterator also move towards the previous direction by $\beta_k$ for momentum. If we choose $\beta_k = \beta$ to be a constant, the scheme yields
$$
\omega^{(k+1)}=\omega^{(k)} - \sum_{j=1}^k \eta_j \beta^{k-j} \nabla f_{i_j}(\omega^{(j)}).
$$
We can see that the Momentum method exploits all the previous direction and uses its weighted average as the approximation to the current gradient. Some experiments indicates this method can improve the performance in cases in which the loss function has a narrow valleys. However, the theoretical convergence analysis still remains an open issue. Moreover, we have to allocate extra memory to keep track of previous $\omega$.

$\bm{Gradient~Averaging}$

The Gradient Averaging method \cite{SGAD} is equivalent to the Momentum mentioned above, if we choose the simple arithmetic average to substitute the weighted average in Momentum. The Gradient Averaging method is proved to have a $O(1/k)$ convergence rate, the same as SGD. Its scheme has form
$$
\omega^{(k+1)}=\omega^{(k)} - \frac{\eta_k}{k}\sum_{j=1}^k \nabla f_{i_j}(\omega^{(j)}).
$$

Note that one do not have to store all the previous gradient term. Instead, one can only keep track of the average of previous $k$ stochastic gradients $G_k = \frac{1}{k}\sum_{j=1}^k \nabla f_{i_j}(\omega^{(j)})$, and use formula $G_{k+1} = \frac{k}{k+1} G_k + \frac{1}{k+1} \nabla f_{i_{k+1}}(\omega^{(k+1)})$ to evaluate $G_{k+1}$ on the run. 

$\bm{Iterate~Averaging}$

The idea behind Momentum and Gradient Averaging is to utilize previous gradient information to determine a better descent direction to accelerate convergence. However, rather than averaging the gradients, the previous iterative points $\omega_k$ can also be taken into account. Some authors use the basic SG iteration but take an average over $\omega_k$ values to give a new algorithm. With a suitable choice of step-sizes, this gives the same asymptotic efficiency as Newton-like second-order SG methods and also leads to increased robustness of the convergence rate to the exact sequence of step sizes \cite{SIAG}. The update scheme reads
$$
\bar{\omega}^{(k+1)}  = \omega^{(k)} -\eta_{k} \nabla f_{i_k}(\omega^{(k)}),
$$
$$
\omega^{(k+1)}  = \frac{1}{k+1}\sum_{j=1}^{k+1} \bar{\omega}^{(j)} .
$$
The Iterate Averaging method uses all the previous iterative points and take its average as the next searching point. It has been proved that under certain assumptions of appropriate step-size, this method enjoys a second-order convergence rate\cite{SIAG}. Even for a fixed step-size strategy, it can also show a great robustness to avoid oscillations. But unfortunately, it is extremely sensible to the initial points: a bad starting point can not only hinder the convergence rate but also cause divergence. Besides, it also requires an extra $O(m)$ memory cost as Momentum.

$\bm{Stochastic~Average~Gradient}$

A typical SAG computes the full gradients at the beginning of iteration, in each following step it chooses a sample's gradient randomly to refresh the full gradients. The update scheme reads
$$
\omega^{(k+1)} =  \omega^{(k)} -\frac{\eta_{k}}{n}\sum_{i=1}^n d_i^{(k)},
$$
where the $i_k$ denotes the randomly chosen index, and
\begin{equation*}
d_i^{(k)} = \left\{
\begin{array} {lcl}
\nabla  f_i(\omega^{(k)}), \quad if i = i_k, \\
d_i^{(k-1)}, \quad otherwise.
\end{array} \right.
\end{equation*}
Like FG method, SAG incorporates a gradient with respect to each training sample, but like SGD, in each iteration it only computes the gradient with respect to a single training example and the cost of the iteration is independent of $n$. In \cite{SAG}, the authors show that the SAG iterations have a linear convergence rate. However SAG have at least 2 drawbacks. First it involves the evaluation of full gradient.Even though it calls for full gradient only once, it is hard to implement in some scenarios. Maybe this difficulty can be resolved by arbitrarily choose the initial gradients for each sample(e.g. zero vectors), but the second weakness makes it completely infeasible in certain scenarios: SAG has an extremely large memory cost $O(np)$ because it has to store previous gradient for each sample. There is no free lunch. These are prices we have to pay for linear convergence.

$\bm{SVRG}$

Stochastic variance reduced gradient(SVRG) introduces an explicit variance reduction method for SGD\cite{SVRG}. SVRG method separates the training process into several epochs, in the beginning of each epoch, it requires the computation of full gradient. And during each epoch, one randomly chooses a sample's gradient to refresh the full gradient. The update scheme reads
$$
\omega^{(k+1)} =  \omega^{(k)} -\eta(\nabla  f_i(\omega^{(k)}) -\nabla  f_i(\tilde{\omega}) + \tilde{\mu} )
$$
where $\tilde{\mu}=\frac{1}{n}\sum_{i=1}^n (\nabla  f_i( \tilde{\omega})$ is the full gradient, $\tilde{\omega}$ is updated at the beginning of each epoch.

SVRG also has a linear convergence rate, but there are at least 3 parameters to tune: the number of epoch, the iteration number in each epoch and the learning rate. Moreover, SVRG has to evaluate full gradient several times, which will also restricts its application in large scale context. Another variation of SVRG is SAGA  \cite{SAGA} which is claimed to support non-strongly convex problems directly and has a better convergence rate. It is essentially at the midpoint between SVRG and SAG. 

$\bm{SCSG}$

Stochastically controlled stochastic gradient(SCSG) is a variation of SVRG. As a great improvement of SVRG, the computation cost and the communication cost of SCSG do not necessarily scale linearly with sample size $n$\cite{SCSG}. 

The main point of SCSG is replacing the full gradient at the beginning of each epoch with batch gradient, and drawing a poisson random number to determine the number of sample to visit in each epoch. However for this algorithm, the batch size is very sensitive. The optimal choice requires some priori knowledge concern to the total loss function.

$\bm{Adagrad}$

Adagrad \cite{adagrad} is an algorithm for gradient-based optimization, it adapts the learning rate to the parameters, performing larger updates for infrequent and smaller updates for frequent parameters. For this reason, it is well-suited for dealing with sparse data. Adagrad can greatly improve the robustness of SGD because it allocates heterogeneous learning rates on different components of $\omega$ at each iteration step. The update scheme reads
$$
\omega^{(k+1)}_i =  \omega^{(k)}_i -\frac{\eta}{\sqrt{G_{k,ii} + \varepsilon
}} \cdot \nabla  f_i(\omega^{(k)}) ,
$$
$G_k \in \mathbb{R}^{p \times p}$ here is a diagonal matrix where each diagonal element is the sum of the squares of the gradients with respect component-$i$ up to iteration step-$k$, and $\varepsilon$ us a smoothing term that avoids division by zero. 

Adagrad’s main weakness is the procedure of accumulating squared gradients in the denominator: Since each added term is positive, the accumulated sum keeps growing during training. This causes the learning rate to shrink and eventually becomes infinitesimally small. As a result, the algorithm is no longer able to acquire additional knowledge. Besides, although Adagrad provides a self-adaptive way to dynamically change the learning rate $\eta$, its initial value still remains to be tuned.

$\bm{Adadelta}$

Adadelta \cite{adadelta} is an extension of Adagrad that seeks to reduce its aggressive, monotonically decreasing learning rate. Instead of accumulating all past squared gradients, Adadelta restricts the window of accumulated past gradients to some fixed size $w$.
Instead of inefficiently storing $w$ previous squared gradients, the sum of gradients is recursively defined as a decaying average of all past squared gradients. The running average $\mathbb{E}[g^2]_t$ at time step $t$ then depends (as a fraction $\gamma$ similarly to the Momentum term) only on the previous average and the current gradient:
\begin{equation} \label{adadelta_gamma0}
\mathbb{E}[g^2]_t =\gamma \mathbb{E}[g^2]_{t-1} + (1-\gamma) g_t^2
\end{equation}
And then they define another exponentially decaying average, this time not of squared gradients but of squared parameter updates:
\begin{equation} \label{adadelta_gamma}
\mathbb{E}[\Delta \theta^2]_t =\gamma \mathbb{E}[\Delta \theta^2]_{t-1} + (1-\gamma) \Delta \theta^2_t
\end{equation}
And the update scheme reads
\begin{equation} \label{adadelta_eps}
\omega^{(k+1)}_i =  \omega^{(k)}_i - \frac{\sqrt{ \mathbb{E}[\Delta \theta^2]_{t-1} + \varepsilon}}{\sqrt{\mathbb{E}[g^2]_t}+\varepsilon } \cdot \nabla  f_i(\omega^{(k)}).
\end{equation}
With Adadelta, we do not need to set a default learning rate since it has been eliminated from the update rule. The weakness of Adadelta, the same as Adagrad, is that it needs extra $O(m)$ memory cost since each component has a different learning rate. 

\section{The Greedy Step Averaging Algorithm}
The greedy step averaging(GSA) algorithm we are proposing is an enhancement of SGD and therefore can serve as a general optimization framework for a variety class of machine learning algorithms. For example, linear regression, logistic regression, softmax regression, etc. The basic idea of the GSA algorithm is to perform exact line search for each sample's loss function step by step. Specifically, consider the general loss function
$$
\text{minimize}_{\omega \in R^p} \quad Loss(\omega) = \sum_{i=1}^{n} l_i(\omega),
$$
where $\omega \in R^p$ denotes the model parameter, $l_i(\omega) = Loss(\omega, x_i, y_i)$ denotes the loss function with respect to $i$-th training instance $(x_i, y_i)$. During a typical iteration step of SGD, the iterator takes a training instance, chooses the negative gradient $-\frac{\partial l_i}{\partial \omega}$ of the local loss function $l_i$ as the descent direction, then calculates an optimal step length $eta_{t,i}$ ensuring $l_i(\omega)$ to reach the minimum. The update scheme of $\omega$ is thus given by
$$
\omega^{(t+1)} = \omega^{(t)} -\eta_{t,i}\frac{\partial l_i}{\partial \omega}.
$$
The step length $\eta_{t,i}$ is calculated through a somehow greedy principle, and that is where the greedy step in the nomenclature of our algorithm comes from. 

\subsection{Linear Regression}
For linear regression, $\eta_{t,i}$ has a closed form. Consider its quadratic loss function
$$
l_i(\omega) = \frac{1}{2} (y_i - \omega^T x_i)^2,
$$
where the intercept term is already absorbed in $x_i$. We can write the update scheme as following:

$$
\omega^{(t+1)} = \omega^{(t)} - \eta_{t,i} x_i (y_i - \omega^{(t)} x_i)
$$
Substitute (1) into $l_i(\omega^{(t+1)}) = 0$ yields 
$$
\eta_{t,i} = \frac{1}{x_i^T x_i}.
$$

\subsection{Logistic Regression}
For logistic regression, $\eta_{t,i}$ should be obtained by tricks. Consider its cross-entropy loss function
$$
 l_i(\omega)= -y_i \beta^T x_i +\log \big(1+\exp(\omega^T x_i)\big) ,
$$
where $p_i = \frac{1}{1+\exp(-\omega^T x_i)}$, $\frac{\partial l_i(\omega)}{\partial \omega} = - x_i(y_i - p_i)$. Thus $l_i(\omega^{(t+1)}) = 0$ yields an intractable solution $\eta_{t,i} \rightarrow \infty$. However, this can be easily overcame by avoiding choosing the target value at sigmoid function's asymptotes. Instead, we set a confidence threshold (in fact this idea is also recommended by Yann Lecun for training neural network \cite{EffBP}) at, for example, $\hat{p}_1=0.95, \hat{p}_0 = 1-\hat{p}_1,$. Each time we see a positive training instance $x_i(y_i = 1)$, we update $\omega$ along the local negative gradient while satisfying $p_i = \hat{p}_1$, and vice versa. That is, 

\begin{equation*}
\left\{
\begin{array} {ll}
p_i = \frac{1}{1+\exp(-\omega^{(t+1)} \cdot x_i)} = \hat{p}_1, \\
\omega^{(t+1)} = \omega^{(t)} + \eta_{t,i} x_i(y_i - p_i).
\end{array} \right.
\end{equation*}

the solution gives
$$
\eta_{t,i} = \frac{\text{sgn}(y_i - 0.5)\log(\hat{p}_1 / \hat{p}_0) - \omega^{(t)} \cdot x_i}{x_i^T x_i(y_i - p_i)}.
$$

\subsection{Softmax Regression}
Softmax regression is an extension of logistic regression which serves as an efficient multi-class classification model. However, the calculation of greedy step length $\eta_{t,i}$ cannot be directly applied to softmax regression model since the solution of $\eta_{t,i}$ does not have a closed form. This problem can be solved by introducing an approximation formula. 
In a softmax regression model, we assign a weight $\omega_k$ to class-$k$ respectively, where $k \in \{1, 2, ..., L\}$. Let $p_i^{(k)} = P(Y = k|x_i) = \frac{\exp{(\omega_kx_i)}}{\sum_{j=1}^L \exp{(\omega_j x_i)}}$ denotes the probability that $x_i$ belongs to class-$k$. Thus the loss function of a training instance $x_i$ with label $k$ is 
\begin{equation}
\begin{array} {ll}
l_i(\omega) &= -\log{\frac{\exp{(\omega_kx_i)}}{\sum_{j=1}^L \exp{(\omega_j x_i)}}}, \\
 &= -\omega_kx_i+\log{(1+\sum_{j=1}^L \exp{(\omega_j x_i)})}.
\end{array} 
\end{equation}
And

\begin{equation*}
 \frac{\partial l_i}{\partial \omega_l} = \left\{
\begin{array} {ll}
-x_i(1 - p_i^{(l)}), l = k, \\
x_ip_i^{(l)}, l \neq k.
\end{array} 
\right.
\end{equation*}

Thus the greedy step length $\eta$ solves the equation(we omit the subscript-$i$ for simplicity):
\begin{equation} \label{eq1}
\begin{array} {ll}
\hat{p}_k &= \frac{\exp{[(\omega_k + \eta (1-p^{(k)})x^T)x]}}{\sum_{j \neq k} \exp{[(\omega_k - \eta p^{(j)}x^T)x]} + \exp{[(\omega_k + \eta (1-p^{(k)})x^T)x]}} \\
&= \frac{\exp(\omega_kx) \cdot \exp((1-p^{(k)}) \eta)} {\sum_{j \neq k} \exp(\omega_jx) \cdot \exp(- p^{(j)} \eta) + \exp(\omega_kx) \cdot \exp((1-p^{(k)}) \eta)}, \\
& \quad (\lambda = \eta x^T x) \\
&= \frac{e^{\lambda} e_k \cdot b_k^{-\lambda}}{\sum_{j \neq k} e_j \cdot b_j^{-\lambda} + e^{\lambda} e_k \cdot b_k^{-\lambda}} \\
&\approx \frac{e^{\lambda} e_k \cdot b_k^{-\lambda}}{\sum_{j = 1}^L e_j \cdot b_j^{-\lambda}}, (\lambda << 1).
\end{array} 
\end{equation}
where $e_j = \exp(\omega_j x), b_j = \exp(p^{(j)}).$ Notice that $b_j \in (1, e), \frac{b_k}{e} \in (e^{-1}, 1)$. Since we assume $\lambda << 1$, thus we can apply linear approximation to the right hand side of (\ref{eq1}):
$$
b_j^{-\lambda} \approx 1 - \lambda \ln b_j  \approx 1 - (b_j -1) \lambda,
$$
$$
(b_k/e)^{-\lambda} = (e/b_k)^{\lambda} \approx 1 + \lambda \ln (e/b_k) \approx 1 + (e/b_k -1) \lambda,
$$
Substitute into (\ref{eq1}) to give a linear equation with respect to $\lambda$:
\begin{equation} \label{eq2}
\hat{p}_k \sum_{j=1}^{L}e_j(1 +\lambda- b_j \lambda) = e_k (1-\lambda+e \lambda/b_k).
\end{equation}

from (\ref{eq2}) we obtain the final approximation formula
\begin{equation} \label{approx_lambda}
\lambda = \frac{-\hat{p}_k \sum_{j=1}^L e_j +e_k}{\hat{p}_k \sum_{j=1}^L e_j(1-b_j) +e_k - ee_k/b_k} .
\end{equation}
And finally the step length is obtained by
\begin{equation} \label{approx_eta}
\eta = \frac{-\hat{p}_k \sum_{j=1}^L e_j+e_k}{\hat{p}_k \sum_{j=1}^L e_j(1-b_j) +e_k - ee_k/b_k} \cdot \frac{1}{x^Tx}
\end{equation}

\subsection{Logistic regression derived from approximation formula (\ref{approx_eta})}

Admittedly, we can directly utilize softmax regression to classify binary-class datasets in spite of a doubled memory cost compared to classic logistic regression. However, we can conduct a transformation to reduce the extra cost. Here we present the specific form of the approximative greedy step length for logistic regression that is equivalent to the previous introduced softmax regression with binary-class. Denote $p_1 = \frac{1}{1+\exp(-\omega \cdot x)}$, $p_0 = 1-p_1$, $b_i = \exp(p_i), i=0,1$, and $\hat{p} = 0.95$ is the confidence threshold. For Softmax regression with binary class, weight $\omega^{SM} = [ \omega_0, \omega_1]^T$, $p_0=\frac{\exp(\omega_0 x)}{\exp(\omega_0 x)+\exp(\omega_1 x)}, p_1 = 1-p_0$. For instance $(x,y)$, gradient 
$$
g^{SM}= x (y-p_1) \cdot \left[
\begin{matrix}
1  \\
 -1
\end{matrix}
\right],
$$
The update scheme for $\omega^{SM}$ is
\begin{equation} \label{eq3}
\begin{array} {ll}
\omega^{(t+1)} &=\omega^{(t)} -\eta g^{SM} \\
&= \left[\begin{matrix}
\omega_0^{(t)} -\eta x(y-p_1)  \\
 \omega_1^{(t)} +\eta x(y-p_1) 
\end{matrix}
\right].
\end{array} 
\end{equation}
Let $\omega^{LR} = \omega_1 - \omega_0$, $g^{LR} = x(p_1-y)$. From (\ref{eq3}) we can derive the update scheme for $\omega^{LR}$
\begin{equation}
\begin{array} {ll}
\omega^{(t+1)} &=\omega^{(t)} -2 \eta x(p_1-y) \\
&=\omega^{(t)} -2 \eta g^{LR}. \\
\end{array} 
\end{equation}
Therefore the greedy step length for logistic regression is $\eta^{LR} = 2\eta$. Next we calculate $\eta$. From (\ref{approx_eta}) we have
\begin{equation}
\eta = \frac{-\hat{p}(e_0+e_1) +e_y}{\hat{p}[e_0(1-b_0) + e_1(1-b_1)] +e_y - ee_y/b_y} \cdot \frac{1}{x^Tx}, 
\end{equation}
where $e_j = \exp(\omega_j x), b_j = \exp(p_j), j=0,1$, $y = 1$ or $0$ represents the label of instance $x$. 

For $y=1$, use relation $e_0/e_1 = p_0/p_1, b_0b_1=e$, we have
\begin{equation}
\begin{array} {ll}
\eta &= \frac{-\hat{p}(e_0/e_1+1) +1}{\hat{p}[e_0/e_1\cdot(1-b_0) + 1-b_1] +1 - e/b_1} \cdot \frac{1}{x^Tx}\\
&=\frac{-\hat{p}(p_0/p_1+1) +1}{\hat{p}[p_0/p_1\cdot(1-b_0) + 1-b_1] +1 - b_0} \cdot \frac{1}{x^Tx}\\
&=\frac{p_1 - \hat{p}}{\hat{p}(1-p_0b_0-p_1b_1)+p_1(1-b_0)} \frac{1}{x^Tx}. \\
\end{array} 
\end{equation}
Similarly, for $y=0$ we have
\begin{equation}
\eta =\frac{p_0 - \hat{p}}{\hat{p}(1-p_0b_0-p_1b_1)+p_0(1-b_1)} \frac{1}{x^Tx}, \\
\end{equation}

Therefore we obtain the greedy step length 
\begin{equation} \label{approx_eta_forLR}
 \eta^{LR} = \left\{
\begin{array} {ll}
\frac{p_1 - \hat{p}}{\hat{p}(1-p_0b_0-p_1b_1)+p_1(1-b_0)} \frac{2}{x^Tx}, \quad label = 1, \\
\frac{p_0 - \hat{p}}{\hat{p}(1-p_0b_0-p_1b_1)+p_0(1-b_1)} \frac{2}{x^Tx}, \quad label = 0.
\end{array} 
\right.
\end{equation}

\subsection{Averaging Scheme}

The intuition of our algorithm is straightforward: exploit the information of each learning instance as much as possible. However, foreseeably, this could bring great oscillation of global loss function. To address this issue, we propose a dynamic adapting strategy. Its idea is dynamically adapting over time by using previous information of greedy step lengths. A basic observation is that the closer we come to global minimum, the smaller the greedy step length is since an increasing proportion of training data is better classified. Therefore, if the greedy step lengths remain steadily in a relatively low interval during a period of iteration, we should know our model is closer to convergence and should thus avoid large learning rates in future iterative steps. In consideration of memory costs, we calculate the arithmetic mean of all previous greedy step length $E [\eta]_t$ as the empirical learning rate:
$$
E [\eta]_t = \frac{1}{t} \sum_{i=1}^t \eta_i.
$$

In consideration of storage costs, we can compute $E [\eta]_t$ on the run using
\begin{equation} \label{mean_eta}
E [\eta]_t =\frac{t-1}{t} E[\eta]_{t-1} + \frac{1}{t} \eta_t.
\end{equation}


For the complete algorithm details see Algorithm \ref{GSA}. And the general framework of GSA see Algorithm \ref{GSA_gen}.

\begin{algorithm}
\caption{GSA algorithm for LR and Softmax}
\label{GSA}     
\begin{algorithmic}[1]       
\REQUIRE       
   Initial parameter $\omega_0$
\FOR {t in  $i \in [0,T]$} 
\STATE Take a Training Sample $(x_t, y_t)$;
\STATE Compute Probability $p_t$, Gradient $g_t$;
\STATE Compute Greedy Step Size $\eta_t=f(x_t, y_t, p_t)$ by (\ref{approx_eta});
\STATE Compute Averaged Greedy Step Size $\bar{\eta}=mean(\eta_t)$;
\STATE Apply Update $\omega_{t+1}=\omega_t-\bar{\eta} g_t$;
\ENDFOR
\end{algorithmic}
\end{algorithm}

\begin{algorithm}
\caption{GSA algorithm in general}
\label{GSA_gen}     
\begin{algorithmic}[1]       
\REQUIRE       
   Initial parameter $\omega_0$, loss function $L(\omega) = \sum_{i=1}^N l_i(\omega)$
\FOR {t in  $i \in [0,T]$} 
\STATE Take a Training Sample $(x_t, y_t)$;
\STATE Compute Stochastic Gradient $g_t = \frac{\partial l_t}{\partial \omega}$;
\STATE Compute Greedy Step Size $\eta_t$ by exact line search on $\l_t(\omega_t - \eta g_t)$;
\STATE Compute Averaged Greedy Step Size $\bar{\eta}=mean(\eta_t)$;
\STATE Apply Update $\omega_{t+1}=\omega_t-\bar{\eta} g_t$;
\ENDFOR
\end{algorithmic}
\end{algorithm}

\section{Convergence Analysis}
Now we establish the convergence theory of GSA. Throughout this section, we suppose $f(\omega)$ is the objective function, which takes the form of
$$ f(\omega)=\frac{1}{n}\sum_{i=1}^{N}L(\omega,x_i), $$
where $L(\omega,x_i)$ is the standard log-loss function, the update scheme of the $k$-th iteration reads
$$ \omega^{(k+1)} = \omega^{(k)} -\eta^{(k)}\nabla f_i(\omega^{(k)}), $$
where $\eta^{(k)}>0$ is the learning rate given by (\ref{approx_eta}) and $i$ is randomly extracted from $\{1,2,...,N\}$. First we deduce some basic observation of $\eta^{(k)}$. From (\ref{approx_eta}) we have
\begin{equation} \label{eta_upperbound}
\begin{array} {ll}
\eta_k &= \frac{-\hat{p} \sum_{j=1}^L e_j+e_k}{\hat{p}_k \sum_{j=1}^L e_j(1-b_j) +e_k - ee_k/b_k} \cdot \frac{1}{x^Tx}  \\
&=\frac{\hat{p}-p_k}{\hat{p} |1 - \bm{p} \cdot \bm{b}| +p_k |1-e/b_k|} \cdot \frac{1}{x^Tx} \\
&\leq \frac{|\hat{p}-p_k|}{ \bm{p} \cdot \bm{b} -1} \cdot \frac{1}{x^Tx} \\
&\leq \frac{|\hat{p}-p_k|}{ |e^{1/L} - 1| \|x\|^2} \\
&\leq C\cdot |\hat{p}-p_k| ,
\end{array}
\end{equation}
where $\bm{b} = (b_1, \cdots, b_L), \bm{p} = (p_1, \cdots, p_L), p_i = e_i / \sum_{j=1}^L e_j$. And the second inequality holds because $\sum_{i=1}^L p_i = 1$ and $\prod_{i=1}^L b_i = e$, $C$ is some constant independent of $k$. Therefore
\begin{equation} \label{eta_upperbound}
\begin{array} {ll}
\eta^{(k)} &= \frac{1}{k} \sum_{i=1}^k \eta_i \\
& \leq \frac{C}{k}\sum_{i=1}^k |\hat{p}-p_i| \\
& \leq C.
\end{array}
\end{equation}
is bounded. Not only $\eta^{(k)}$ is bounded, we can also assume $ \lim_{k \rightarrow \infty} \eta^{(k)} = \eta_0$ since in each pass $|\hat{p}-p_i|$ is almost sampled from an identical distribution.

On the other hand, we have the denominator of (\ref{esti_Ef})
\begin{equation} \label{eq4}
\begin{array} {ll}
\sum_{k=1}^{T}\eta^{(k)} &=\sum_{k=1}^{T} \frac{1}{k} \sum_{i=1}^k \eta_i \\
&=\sum_{i=1}^{T} \eta_i  \sum_{k=i}^T  \frac{1}{k} \\
&\geq \eta_1 \sum_{k=i}^T  \frac{1}{k} \\
& \rightarrow \infty, \quad \text{as} \quad T \rightarrow \infty.
\end{array}
\end{equation}

Now we present our conclusion. Let $\omega^{*}$ stands for the minimizer of $f$.
Suppose
$$\exists M>0, s.t. \forall k, \mathbb{E}[||\nabla f_i(\omega^{(k)})||^2] \leq M,$$
$$\exists G>0, s.t. \mathbb{E}[||\omega^{(0)}-\omega^{*}||] \leq G,$$
define
$$f_m(T)=min\{f(\omega^{(0)}), f(\omega^{(1)}), ..., f(\omega^{(T)})\},$$
then we claim that
$$\mathbb{E} [f_m(T)] \rightarrow f(\omega^{*}) + O(\eta_0),\quad as \quad T \rightarrow \infty.$$
Proof.
Taylor expansion gives
\begin{align*} ||\omega^{(k+1)}-\omega^{*}||^2 &= ||\omega^{(k)}-\eta^{(k)}\nabla f_i(\omega^{(k)})-\omega^{*}||^2 \\  &= ||\omega^{(k)}-\omega^{*}||^2 -2\eta^{(k)}\nabla f_i(\omega^{(k)})^{T}(\omega^{(k)}-\omega^{*})+[\eta^{(k)}]^2||\nabla f_i(\omega^{(k)})||^2. \end{align*}
Take conditional expectation we obtain
\begin{align*} \mathbb{E}[||\omega^{(k+1)}-\omega^{*}||^2|\omega^{(k)}] &= \mathbb{E}[||\omega^{(k)}-\omega^{*}||^2|\omega^{(k)}] -2 \eta^{(k)}\mathbb{E}[\nabla f_i(\omega^{(k)})^{T}(\omega^{(k)}-\omega^{*})|\omega^{(k)}] \\ &~~~+ [\eta^{(k)}]^2 \mathbb{E}[||\nabla f_i(\omega^{(k)})||^2|\omega^{(k)}] \\ &= ||\omega^{(k)}-\omega^{*}||^2 -2\eta^{(k)}\nabla f(\omega^{(k)})^{T}(\omega^{(k)}-\omega^{*})+ [\eta^{(k)}]^2 \mathbb{E}[||\nabla f_i(\omega^{(k)})||^2|\omega^{(k)}] \\ & \leq ||\omega^{(k)}-\omega^{*}||^2 -2\eta^{(k)}( f(\omega^{(k)}) -f(\omega^{*}))+ [\eta^{(k)}]^2 M. \end{align*}
Take expedition w.r.t $\omega^{(k)}$ and use iteration, the inequality yields
\begin{align*} \mathbb{E}[||\omega^{(k+1)}-\omega^{*}||^2] & \leq \mathbb{E}[||\omega^{(k)}-\omega^{*}||^2] -2\eta^{(k)}( \mathbb{E}[f(\omega^{(k)})] -f(\omega^{*}))+ [\eta^{(k)}]^2 M \\ & \leq ... \\ & \leq \mathbb{E}[||\omega^{(0)}-\omega^{*}||^2] -2\sum_{k=1}^{T}\eta^{(k)}( \mathbb{E}[f(\omega^{(k)})] -f(\omega^{*}))+ M \sum_{k=0}^{T}[\eta^{(k)}]^2 \\ & \leq G -2\sum_{k=1}^{T}\eta^{(k)}( \mathbb{E}[f(\omega^{(k)})] -f(\omega^{*}))+ M \sum_{k=0}^{T}[\eta^{(k)}]^2. \end{align*}
Therefore
$$ 2\sum_{k=1}^{T}\eta^{(k)}( \mathbb{E}[f(\omega^{(k)})] -f(\omega^{*})) \leq G + M\sum_{k=0}^{T}[\eta^{(k)}]^2, $$
Thus
\begin{equation} \label{esti_Ef}
\mathbb{E} [f_m(T)] - f(\omega^{*}) \leq \frac{G + M\sum_{k=0}^{T}[\eta^{(k)}]^2}{2\sum_{k=1}^{T}\eta^{(k)}}. 
\end{equation}

Under our assumption and (\ref{eq4}), we are able to derive an error bound from (\ref{esti_Ef})
\begin{equation} 
\begin{array} {ll}
\lim_{T\rightarrow \infty} (\mathbb{E} [f_m(T)] - f(\omega^{*}))  &\leq \lim_{T\rightarrow \infty}  \frac{G}{2\sum_{k=1}^{T}\eta^{(k)}} + \lim_{T\rightarrow \infty}  \frac{M}{2} \frac{\sum_{k=0}^{T}[\eta^{(k)}]^2}{\sum_{k=1}^{T}\eta^{(k)}} \\
&= 0 + \frac{M}{2} \cdot \lim_{T\rightarrow \infty} \frac{\sum_{k=0}^{T}[\eta^{(k)}]^2}{\sum_{k=1}^{T}\eta^{(k)}} \\
&=\frac{M\eta_0}{2}.
\end{array}
\end{equation}
Thus we established an error bound of $\mathbb{E} [f_m(T)]$ with tolerance $O(\eta_0)$. Similarly, in \cite{SGDTheo1} the author also established an inequality 
$$\|\nabla f(\omega)\| \leq C \eta_0$$ under some reasonable assumptions. According to their arguments, this result is as far as we are able to attain for SGD with learning rate bounded away from zero.

\section{Numerical Experiment}
In this section we make comparisons between GSA and some other state of the art algorithms on several open datasets from libsvm. All the 16 datasets we use here are downloaded from \cite{Libsvm}. The algorithms that we choose as contrasts are SGD, Adadelta, SCSG for the following reason: SGD is the widely accepted classic benchmark, Adadelta and SCSG are representatives of the family of self-adaptive and variance reduction method, respectively. We perform experiments on 16 datasets, and on 15 out of which the gap between average precision score of GSA and the best performance is less than $1\%$(except on letter.scale, GSA is beaten by SCSG by $3.5\%$ in precision after 10 epochs). Due to the space limitations, we only offer the experimental illustrations for 4 of them(w1a, mnist, news20, aloi) and left the results on other 12 datasets in appendix. For those datasets which have corresponding test sets(w1a and news20), we take them for validation while for the other datasets we randomly take $80\%$ from the whole dataset as training set and remain the left $20\%$ for testing. Details of datasets are presented in table \ref{table:3}. We implement GSA for logistic and softmax regression, compare its performance with that of SGD, Adadelta and SCSG. In each experiment, we let the iterator run up to several passes(5, 10, or 20, depend on the required number of passes for loss to saturate) of the training set, and evaluate the log of standard cross-entropy loss, the precision score and the roc-auc score over the whole test set. For each algorithm, the hyper parameters are specified in the legend, and no regularization term is attached. The detailed discussion of the comparison is presented in the following subsections. 
The demonstration of all 3 indicators(loss, precision and roc\_auc score) after 1, 2 and the last pass of training sets is shown in appendix, and we highlights the best score in each columns(we did not test SCSG on the largest dataset url for the limitation of computational resource). For each indicator, we calculate the statistical result between GSA and the best score(the mean of Err = $score_{GSA} - score_{best}$) of other 3 methods over 16 datasets. We also count the time that GSA ranks first in terms of different metrics. The corresponding results are shown in table \ref{table:4}. Compared to the best performances among other 3 methods, GSA is not only robust but also fairly competitive.  

\begin{table}
 \centering\small
 \caption{Comparison of average performance}\label{table:4}
\begin{tabular}{|c|c|c|c|c|c|c|c|c|c|c|}
\hline
 \multirow{2}{*}{problem} & \multirow{2}{*}{metric} & \multicolumn{3}{|c|}{loss} & \multicolumn{3}{|c|}{prec.} & \multicolumn{3}{|c|}{auc}\\ \cline{3-11}
&&  1 & 2 & last & 1 & 2 & last & 1 & 2 & last\\ \hline
\multirow{2}{*}{LR}  
& mean(Err) & 0.056 & 0.016 &  0.019 & -0.016 & -0.018 & -0.011 & -0.009 & -0.005 & -0.001  \\
& \#best out of 7 & 0 & 0 &  1 & 0 & 1 & 0 & 3 & 2 & 3  \\
\hline
\multirow{2}{*}{Softmax}  
& mean(Err) & 0.026 & 0.030 &  0.038 &  0.004 &  -0.004 & -0.001  & / & / & / \\
& \#best out of 9& 2 & 2 &  3 &  3 &  3 & 5  & / & / & / \\
\hline
\end{tabular}
\end{table}

\begin{table}
 \centering\small
 \caption{Dataset information}\label{table:3}
\begin{tabular}{c|c|c|c}
\hline dataset &\#instance(train/test) & \#feature & \#class\\ 
\hline 
 w1a &  2477/47272 & 300  & 2 \\ 
 mnist.scale & 60000 & 780  & 10\\  
 news20.scale & 15935/3993 & 62061 & 20 \\ 
 aloi.scale & 108000 & 128  & 1000\\ 
 \hline

 a9a &  32561 & 123  & 2 \\ 
 breast-cancer\_scale & 683 & 10  & 2\\  
 gistte\_scale & 6000/1000 & 5000 & 2 \\ 
 madelon & 2000/600 & 500  & 2\\ 
 cod-rna & 59535 & 8 & 2\\ 
 url & 2,396,130 & 3231961  &  2\\ \hline
 letter.scale & 15000/5000 & 16  & 26\\ 
 dna.scale & 2000/1186 &  180 & 3\\ 
 sector.scale & 6412/3207 & 55197  & 105\\ 
 usps &7291/2007  & 256  & 10\\ 
 protein & 17766	/6621 & 357  & 3\\ 
 rcv1.multiclass & 15564/518571 & 47236 & 53\\
  \hline
\end{tabular} 
\end{table}

\subsection{Comparison with SGD}
\begin{figure}[H]
\centering
\begin{minipage}[t]{0.41\linewidth} 
\includegraphics[trim=0 0 0 0,scale=0.5,width=\columnwidth]{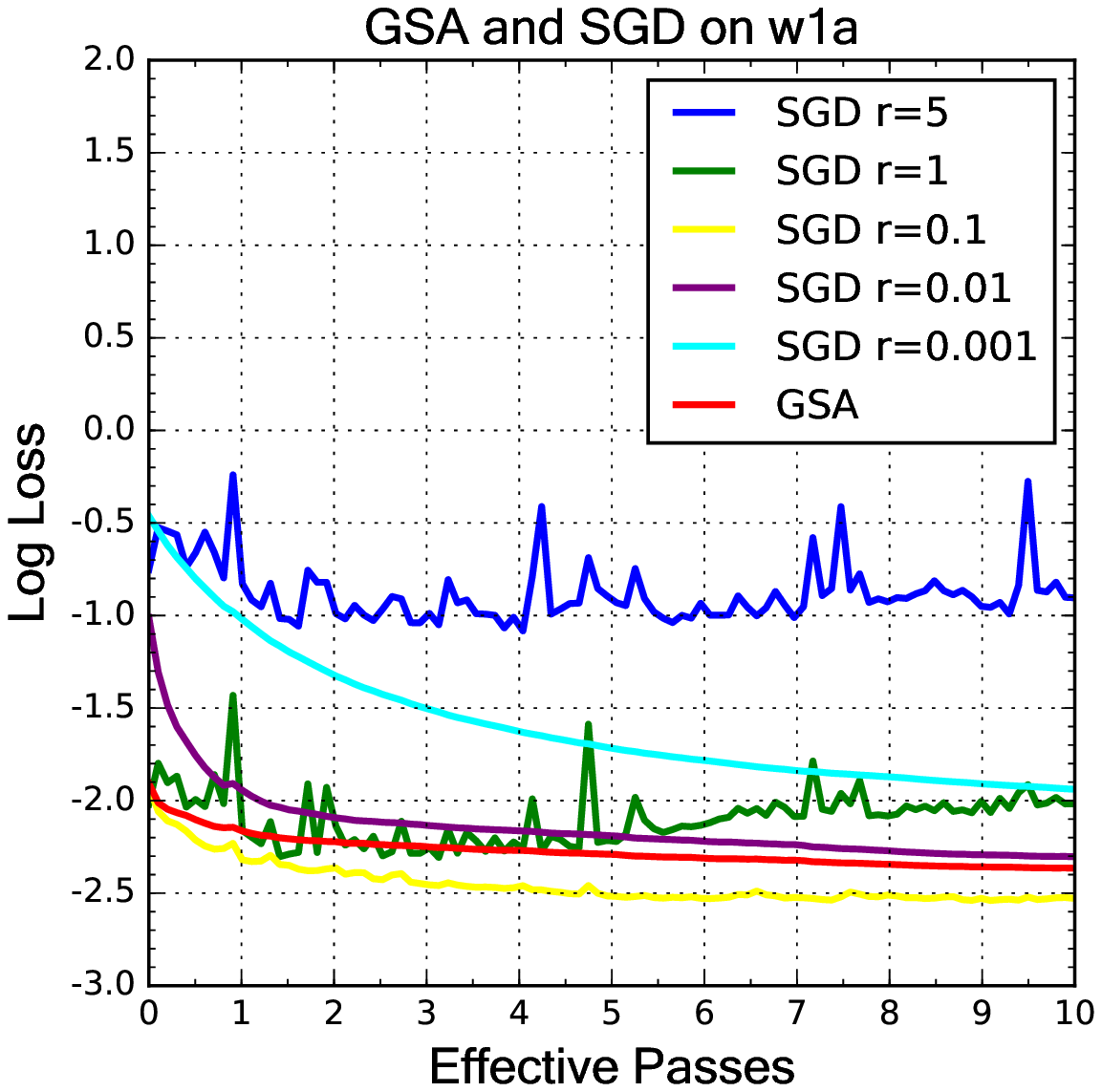} 
\end{minipage} 
\begin{minipage}[t]{0.41\linewidth} 
\includegraphics[trim=0 0 0 0,scale=0.5,width=\columnwidth]{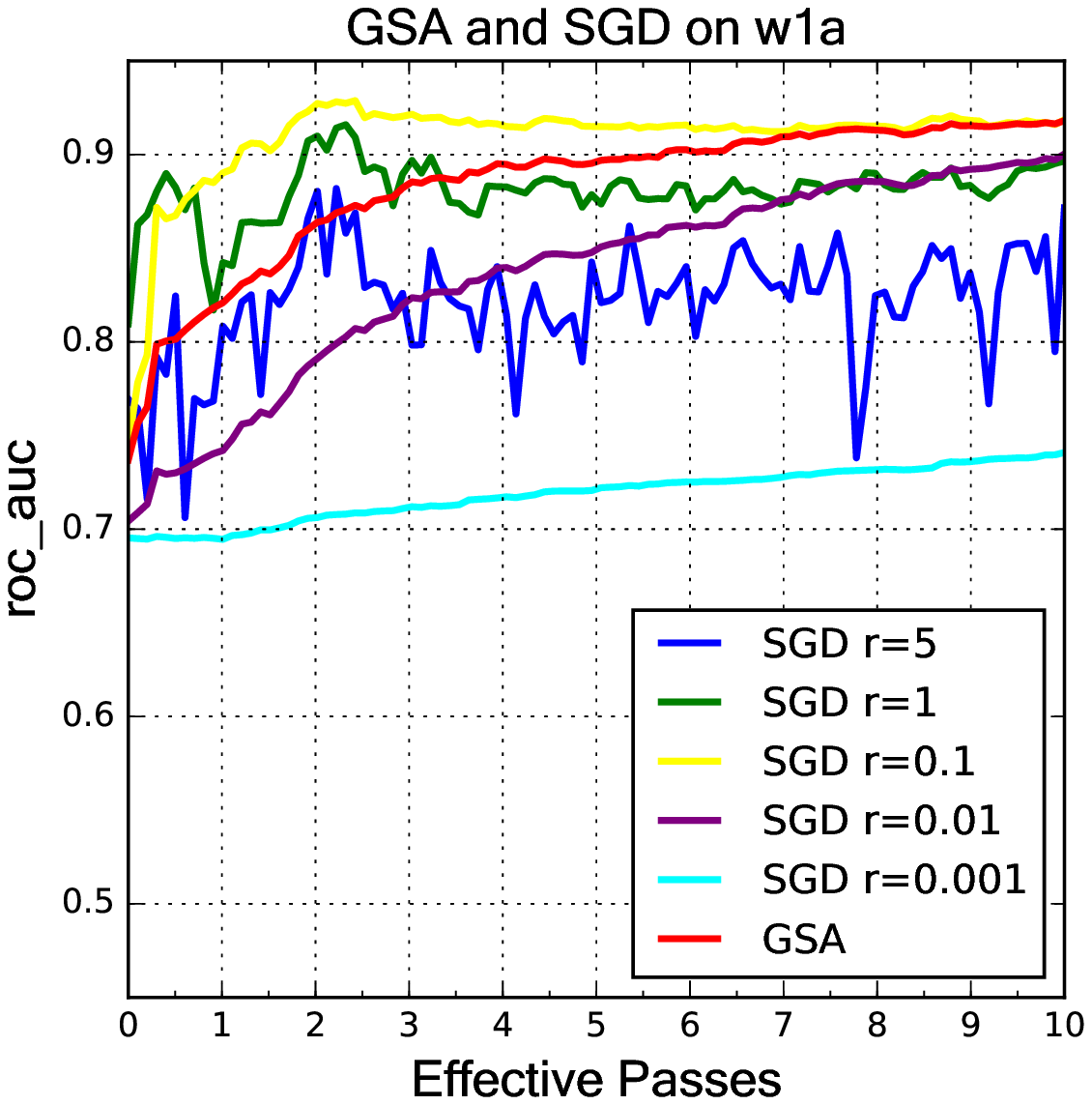} 
\end{minipage} 

\begin{minipage}[t]{0.41\linewidth} 
\includegraphics[trim=0 0 0 0,scale=0.5,width=\columnwidth]{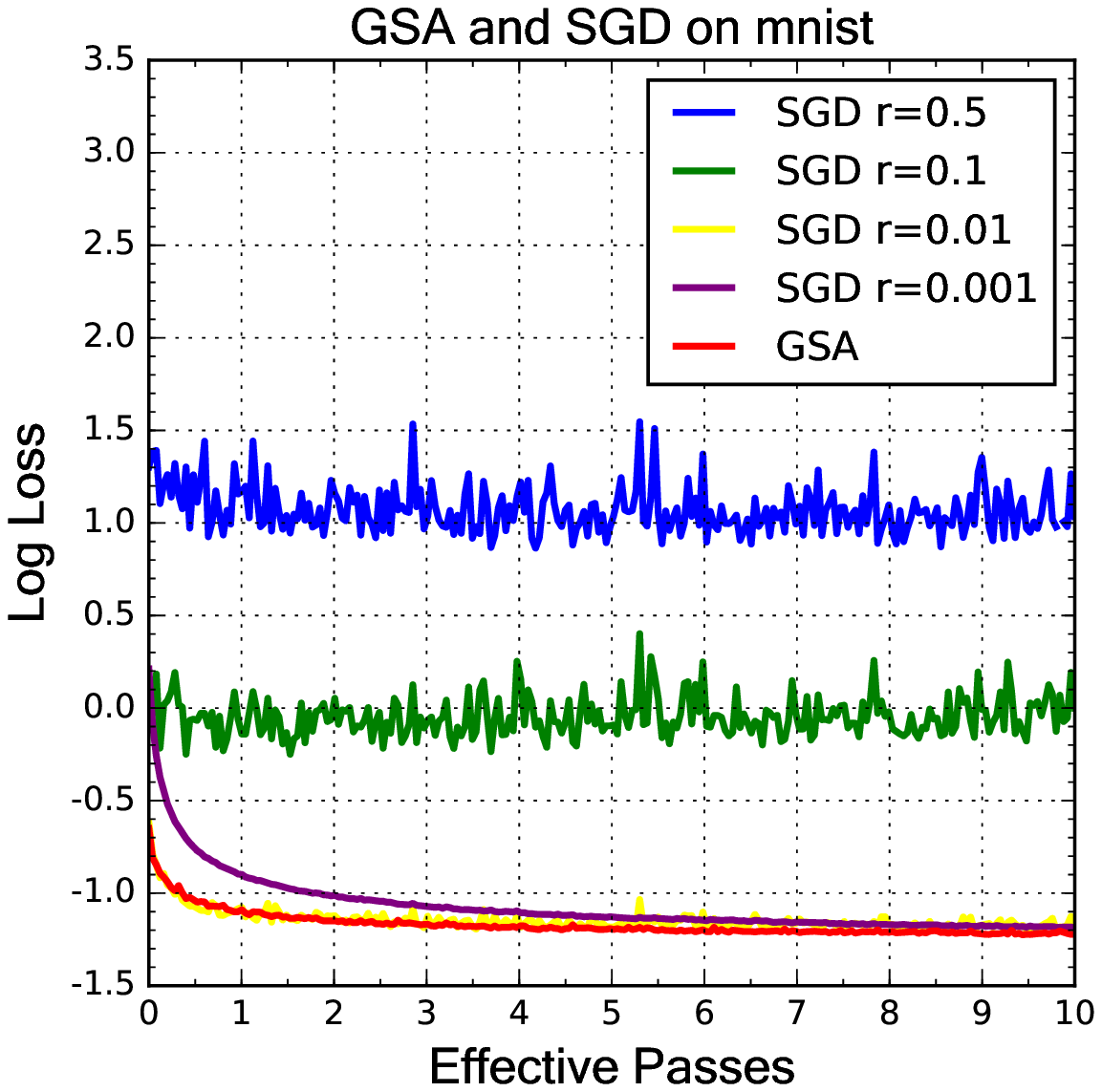} 
\end{minipage} 
\begin{minipage}[t]{0.41\linewidth} 
\includegraphics[trim=0 0 0 0,scale=0.5,width=\columnwidth]{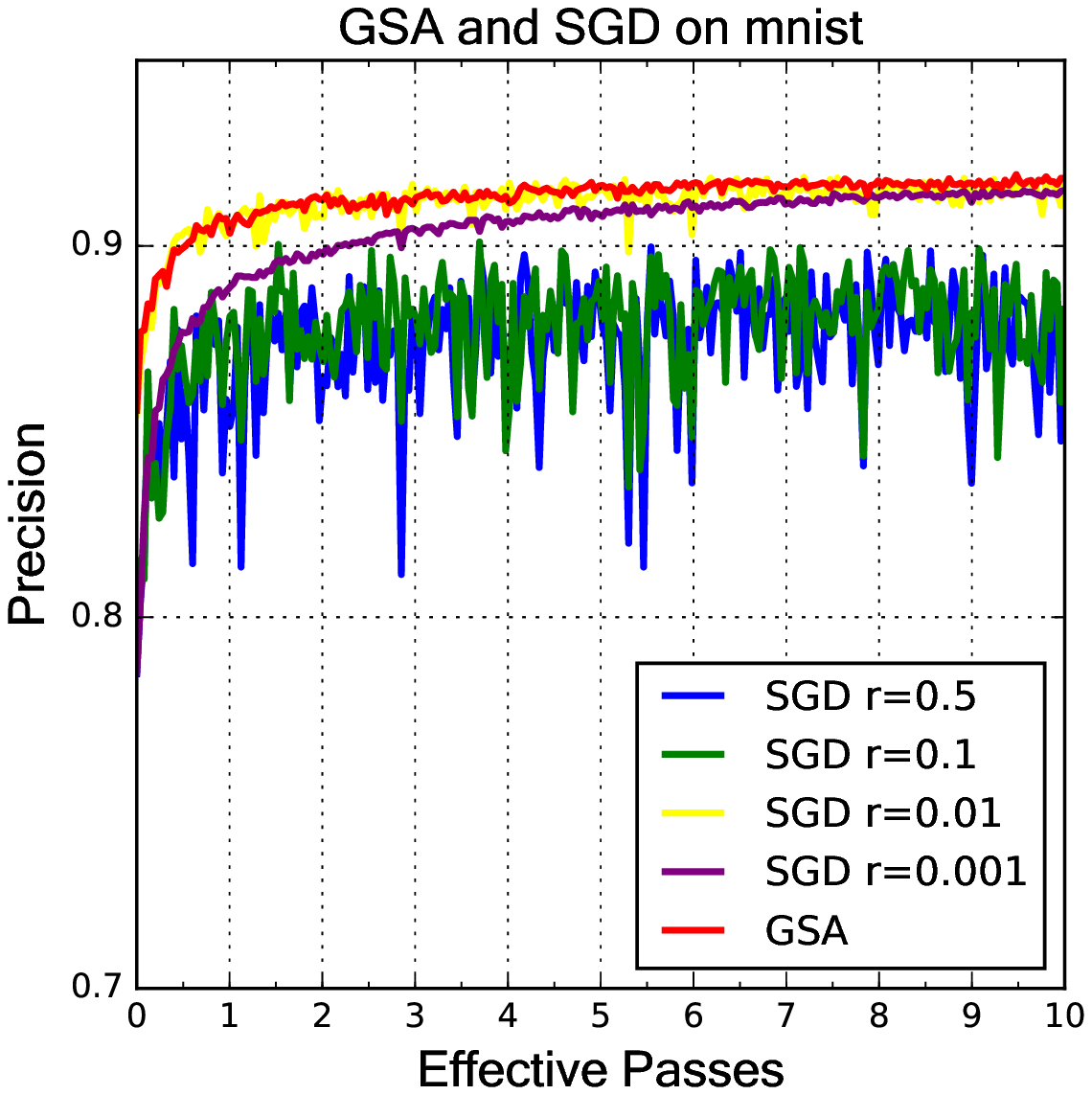} 
\end{minipage} 

\begin{minipage}[t]{0.41\linewidth} 
\includegraphics[trim=0 0 0 0,scale=0.5,width=\columnwidth]{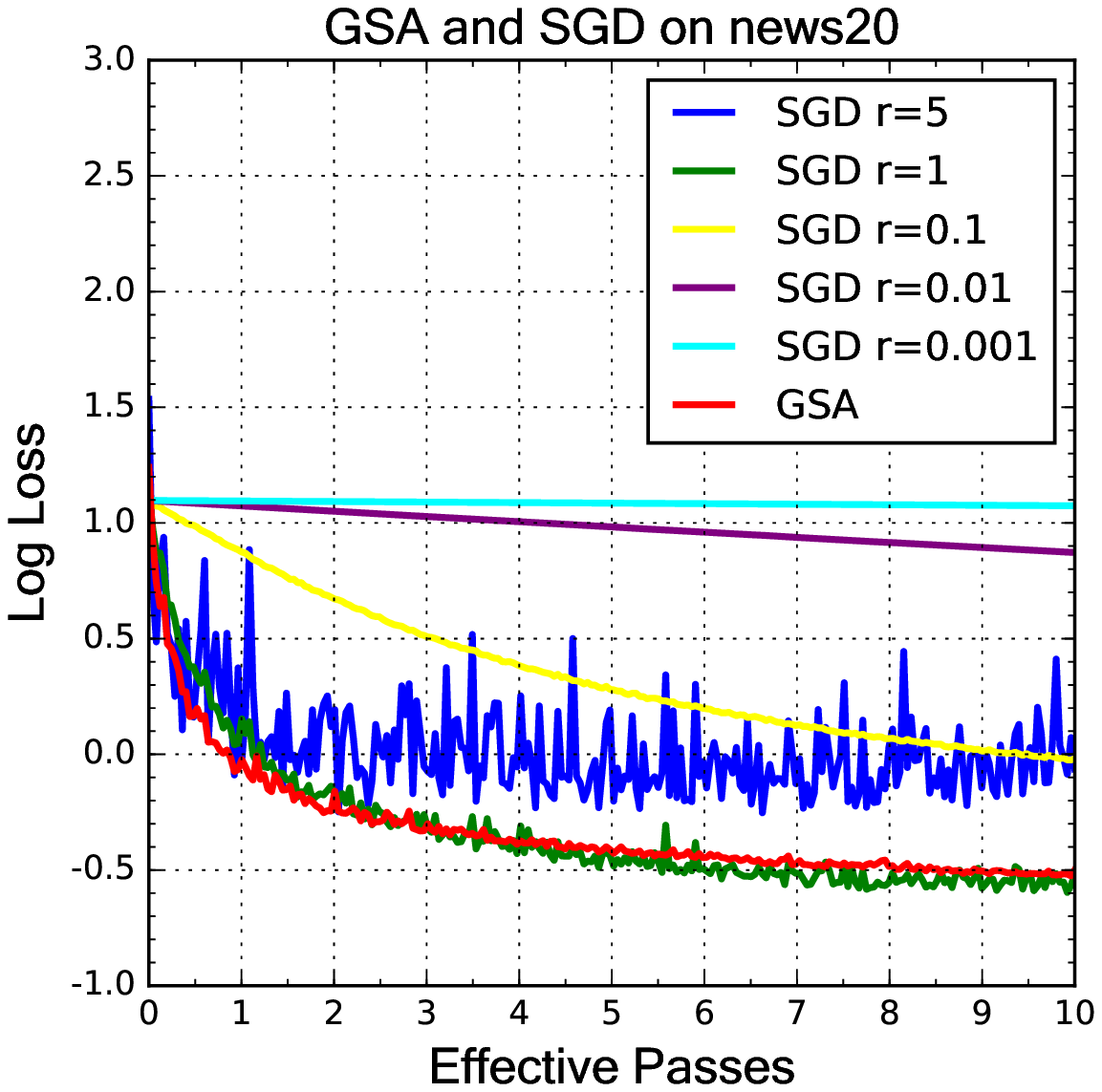} 
\end{minipage} 
\begin{minipage}[t]{0.41\linewidth} 
\includegraphics[trim=0 0 0 0,scale=0.5,width=\columnwidth]{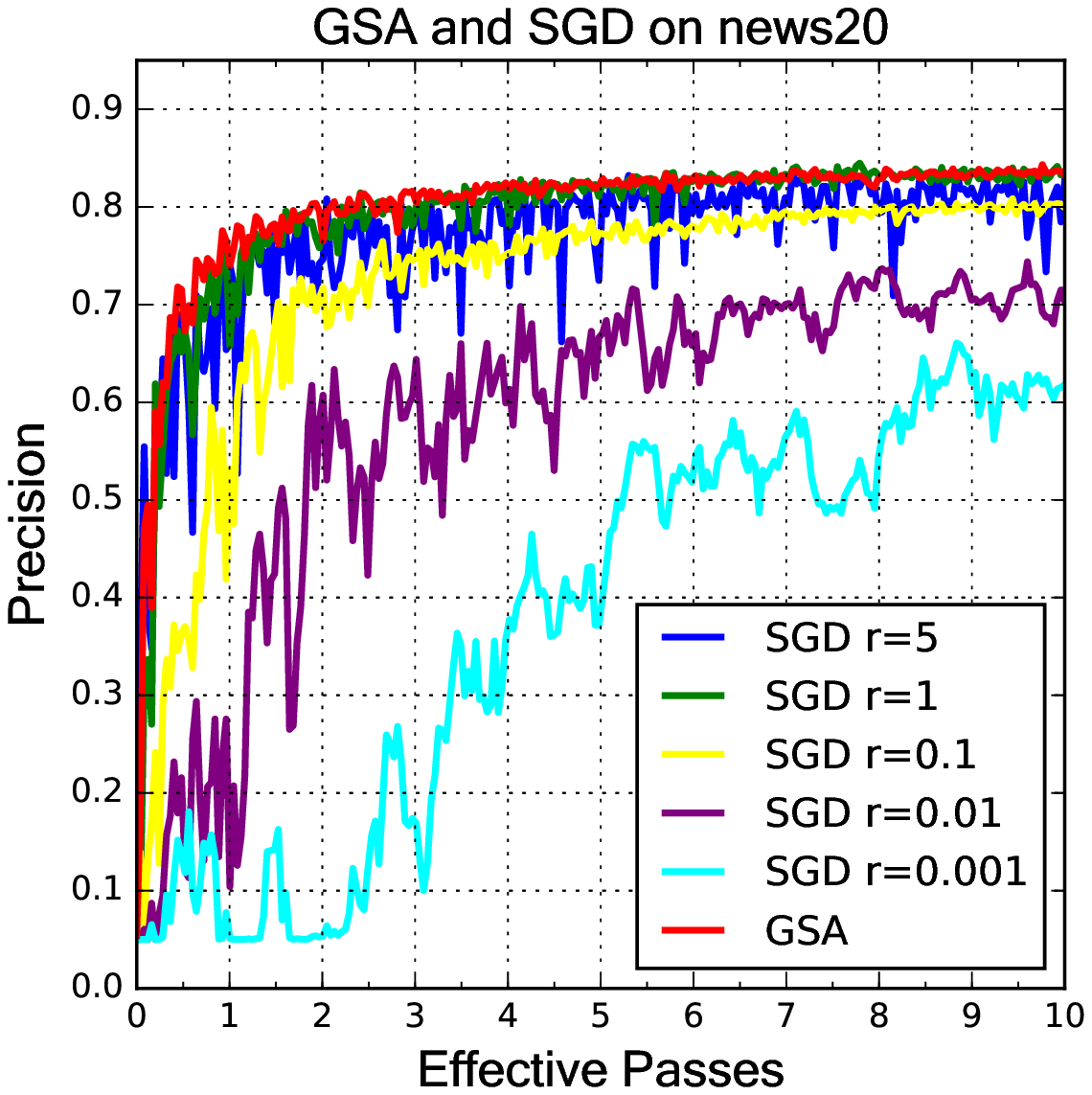} 
\end{minipage} 

\begin{minipage}[t]{0.41\linewidth} 
\includegraphics[trim=0 0 0 0,scale=0.5,width=\columnwidth]{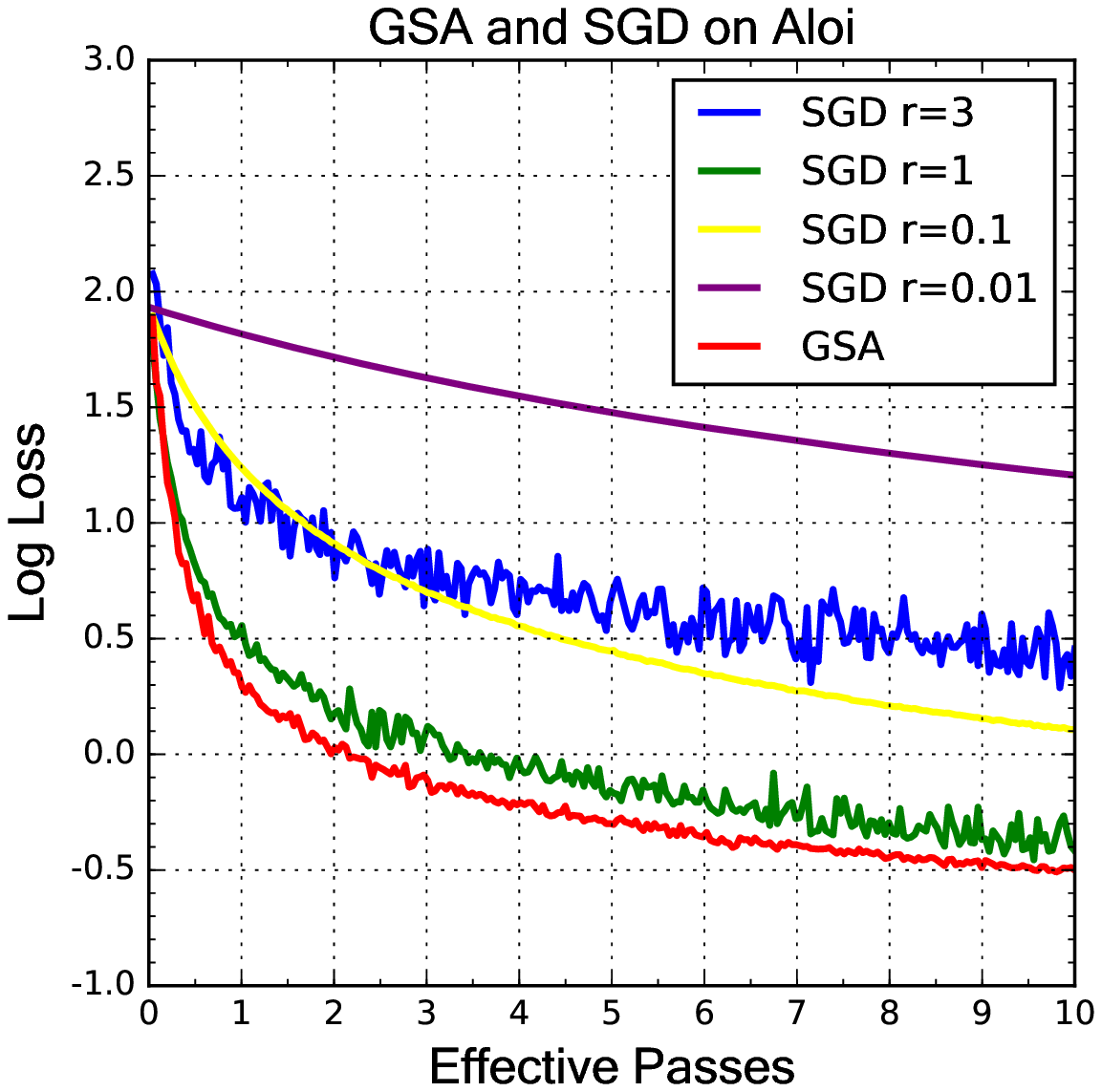} 
\end{minipage} 
\begin{minipage}[t]{0.41\linewidth} 
\includegraphics[trim=0 0 0 0,scale=0.5,width=\columnwidth]{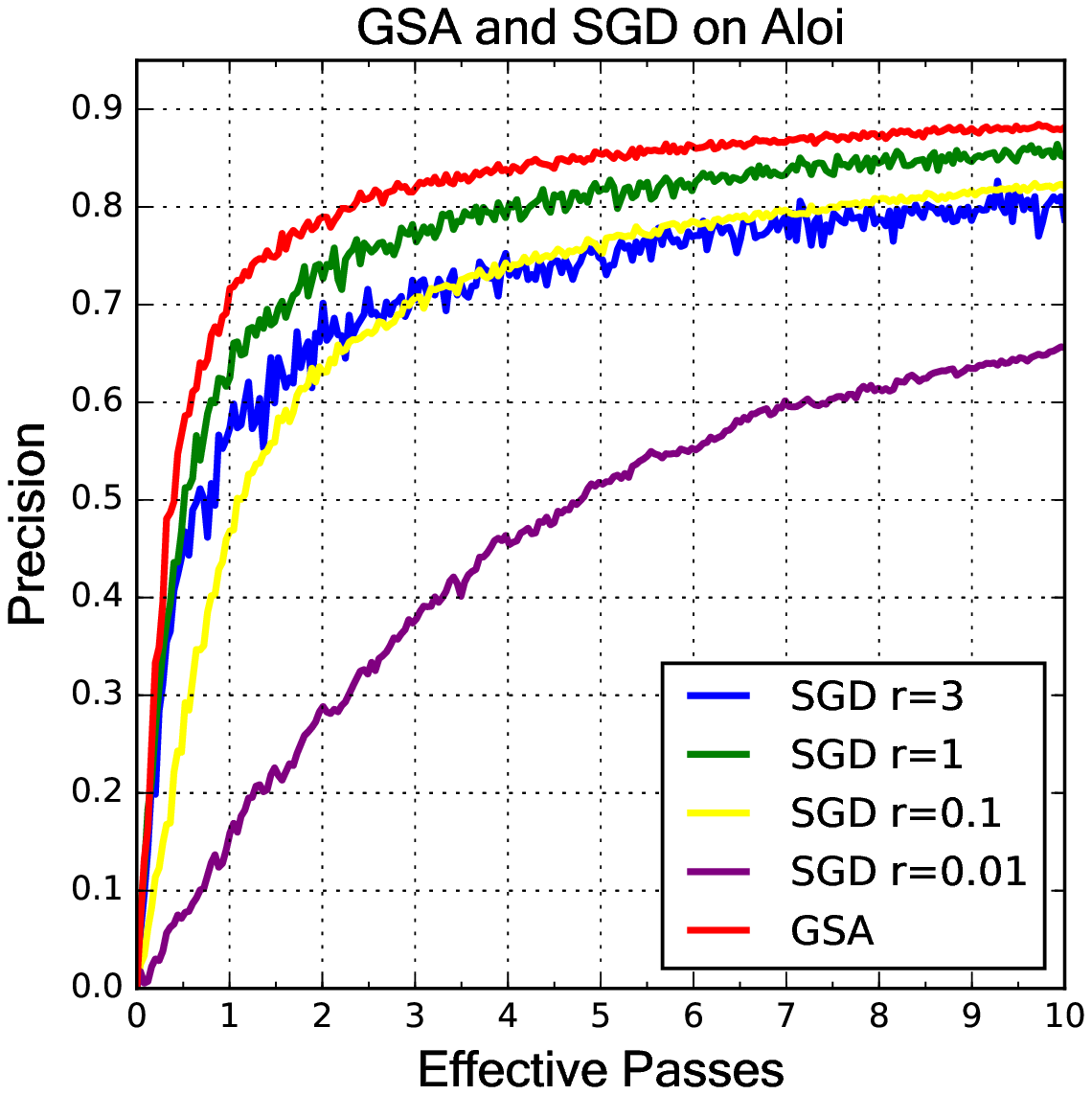} 
\end{minipage} 

\caption{\label{fig2}The performance of GSA and SGD}
\end{figure}

From Fig.\ref{fig2} we can see that the optimal learning rate for SGD varies on different datasets, and GSA shows a better performance than the best behaviour of SGD on mnist, news20 and aloi. That is because GSA is able to capture the magnitude of the optimal learning rate in the early stage of iteration and dynamically change it to adapt to the situation. In another words, GSA can figure out when to stop. Moreover, GSA can also avoid oscillation by detecting the margin from convergence and adjust learning rate correspondingly.    

\subsection{Comparison with Adadelta}
\begin{figure}[H]
\centering
\begin{minipage}[t]{0.41\linewidth} 
\includegraphics[trim=0 0 0 0,scale=0.5,width=\columnwidth]{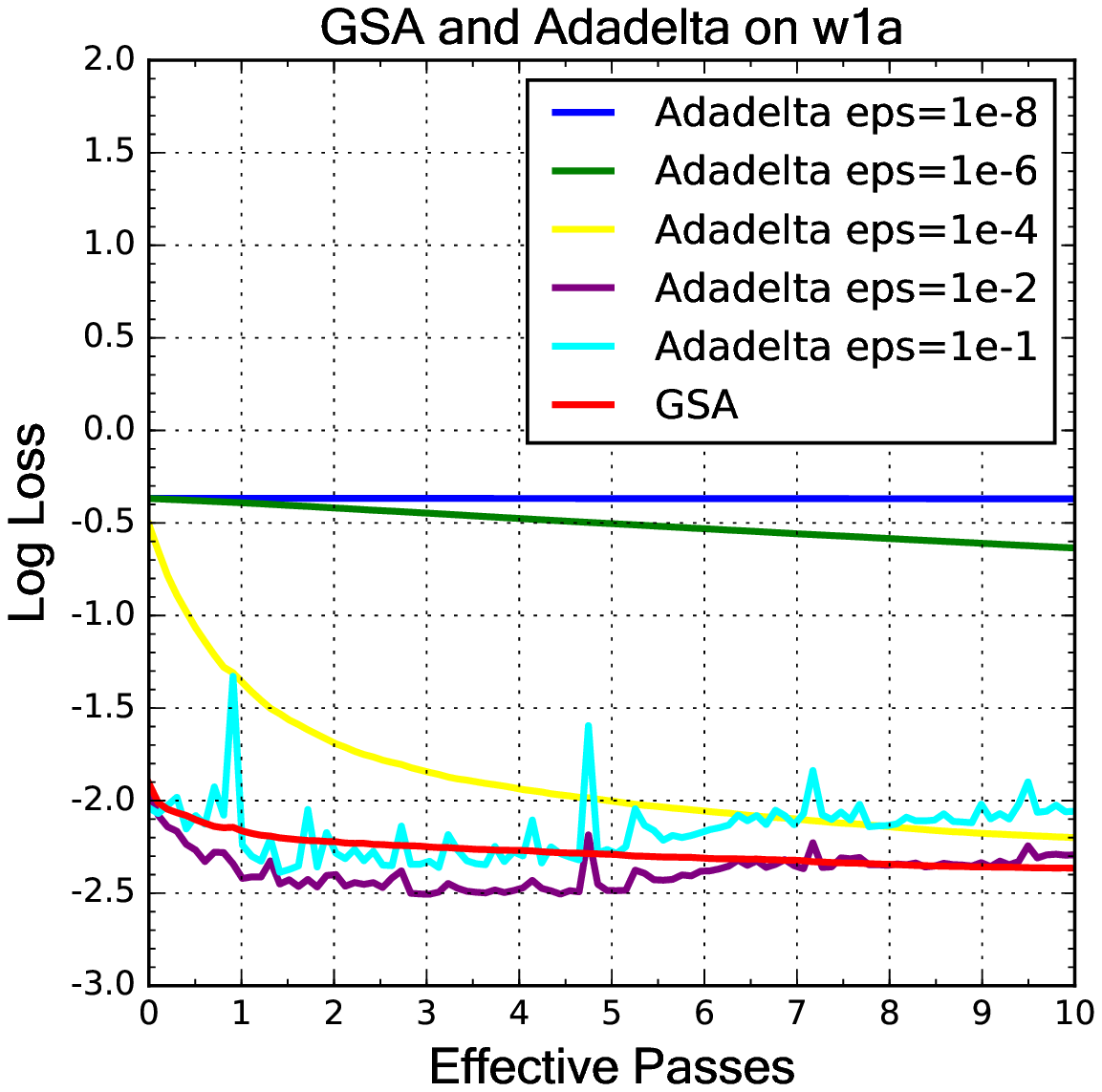} 
\end{minipage} 
\begin{minipage}[t]{0.41\linewidth} 
\includegraphics[trim=0 0 0 0,scale=0.5,width=\columnwidth]{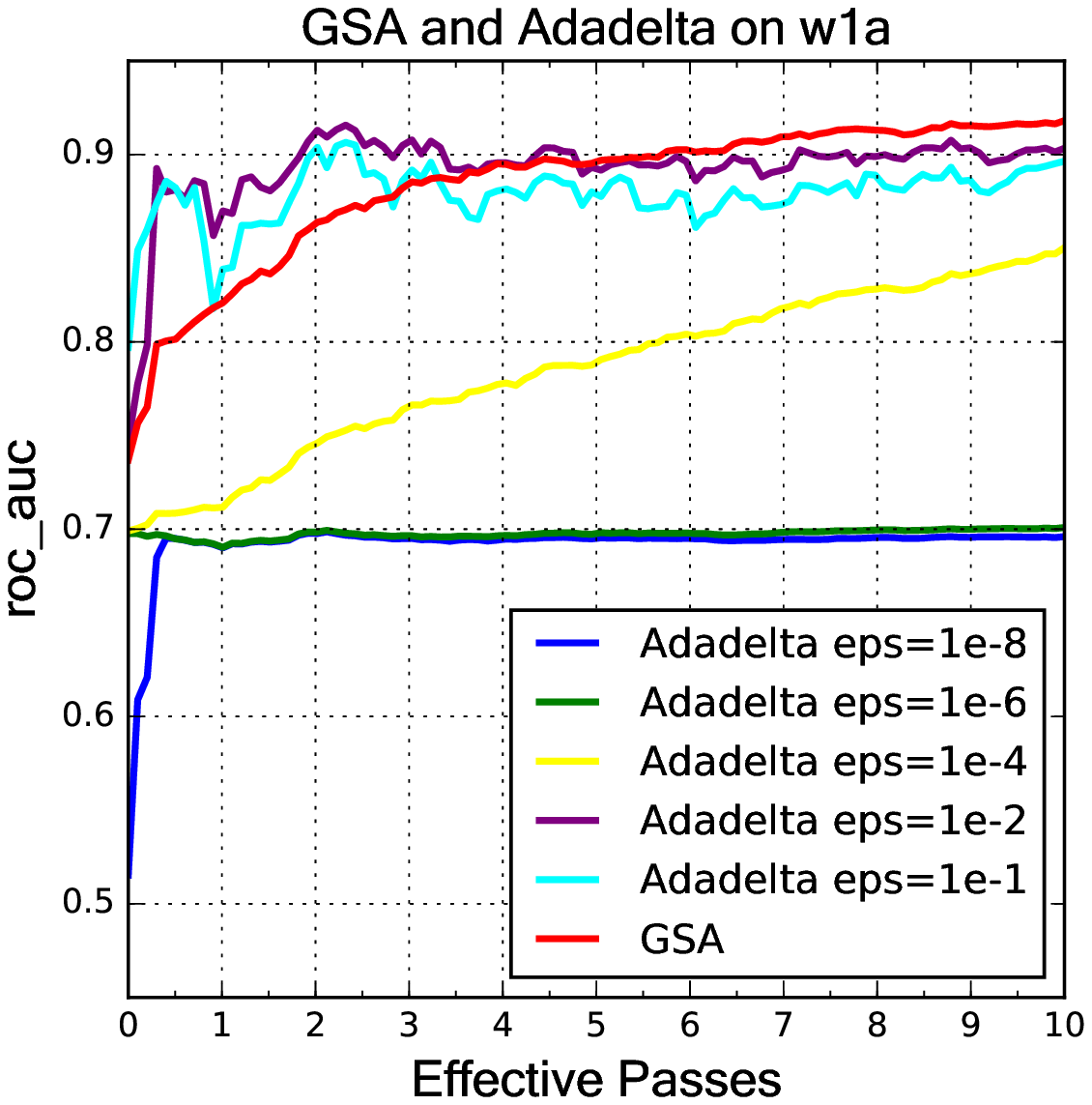} 
\end{minipage} 

\begin{minipage}[t]{0.41\linewidth} 
\includegraphics[trim=0 0 0 0,scale=0.5,width=\columnwidth]{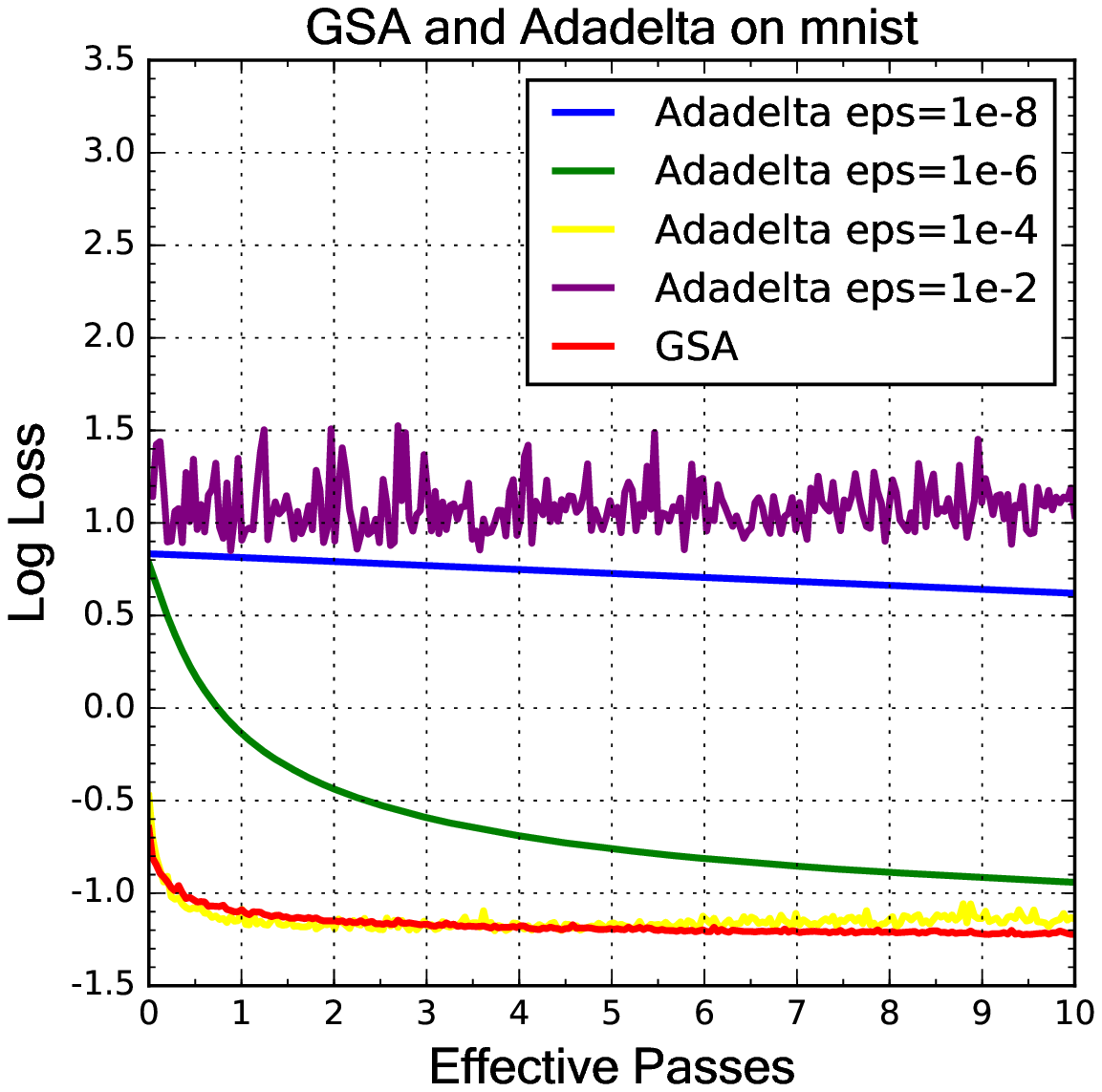} 
\end{minipage} 
\begin{minipage}[t]{0.41\linewidth} 
\includegraphics[trim=0 0 0 0,scale=0.5,width=\columnwidth]{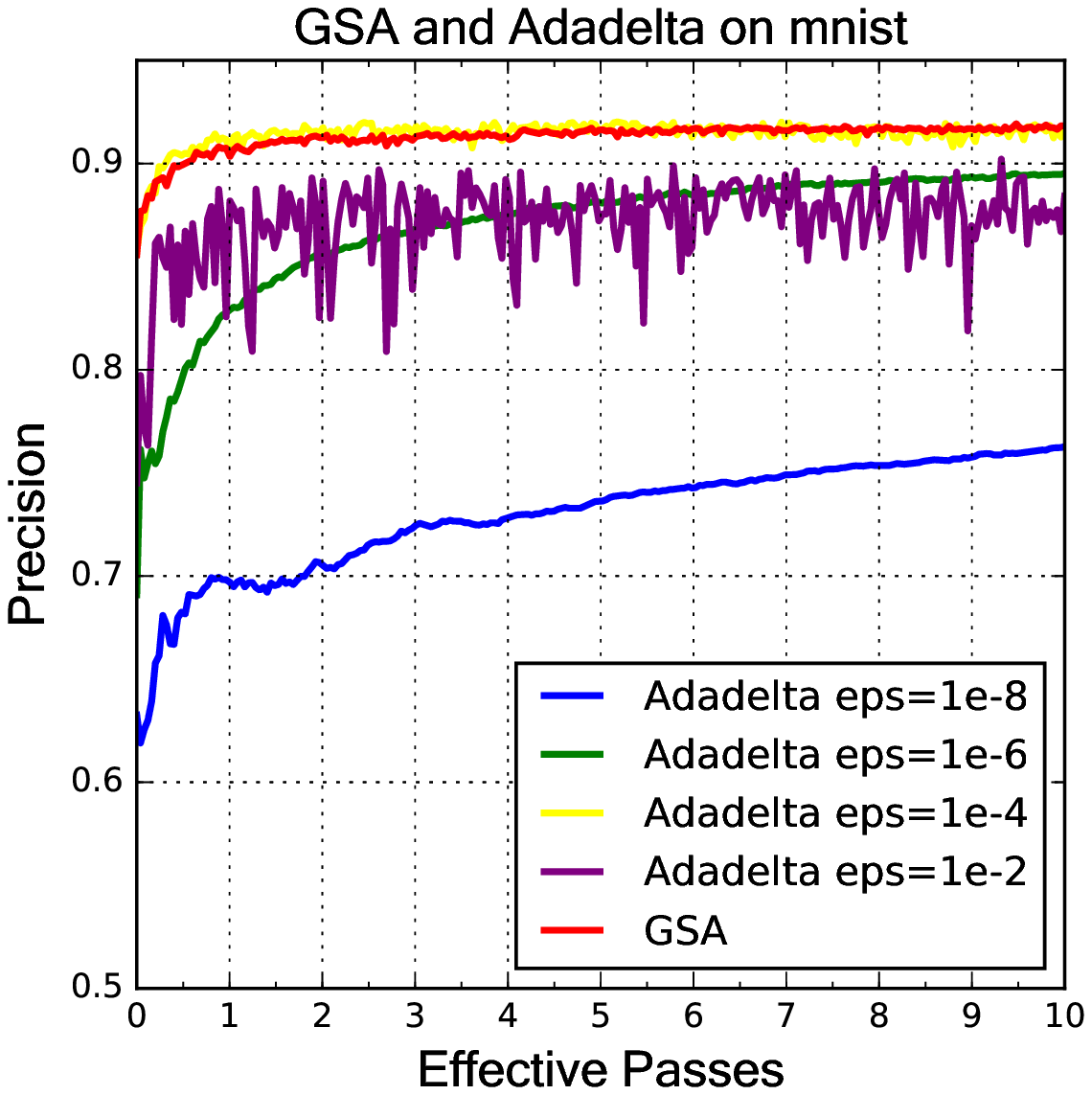} 
\end{minipage} 

\begin{minipage}[t]{0.41\linewidth} 
\includegraphics[trim=0 0 0 0,scale=0.5,width=\columnwidth]{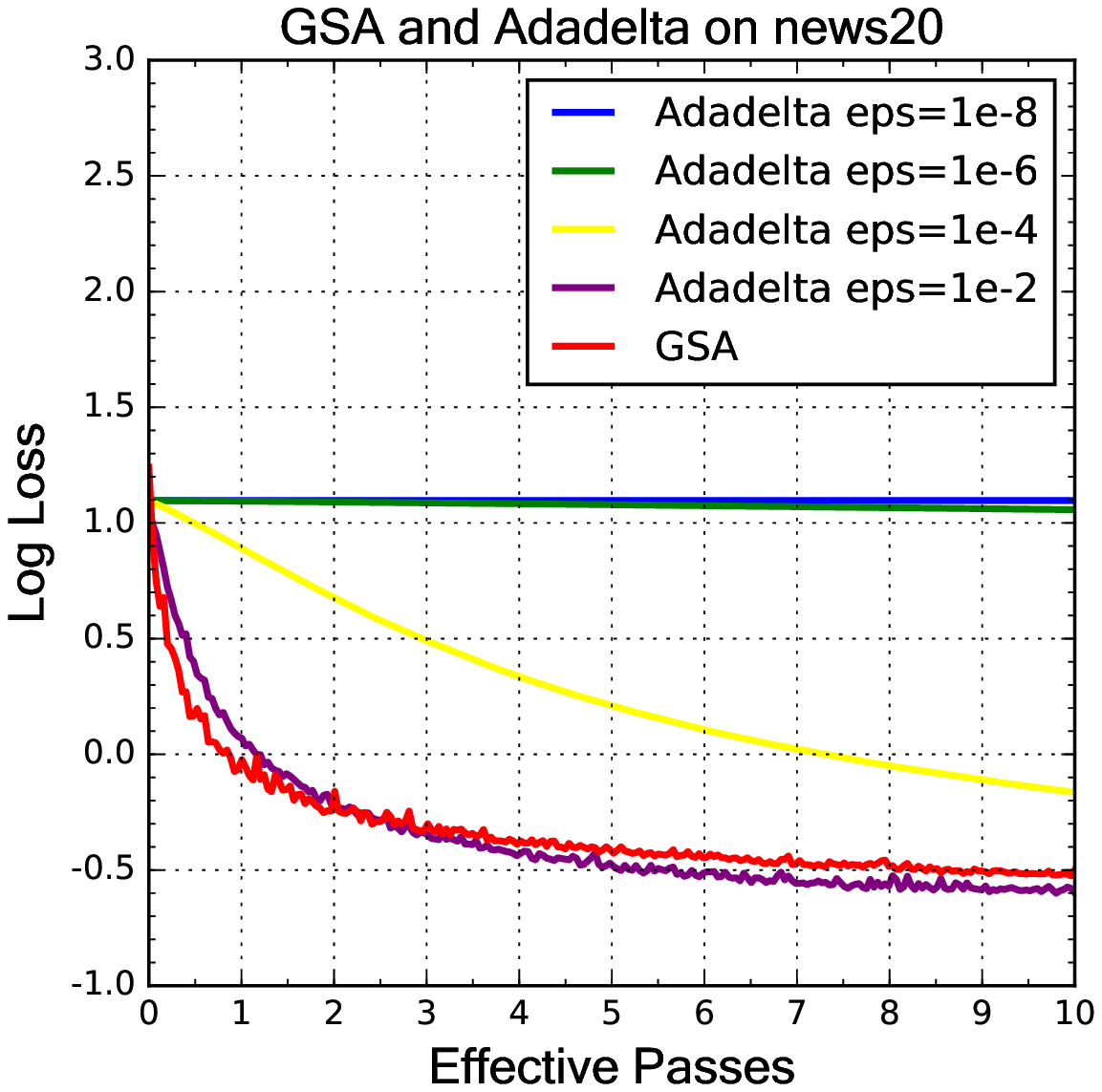} 
\end{minipage} 
\begin{minipage}[t]{0.41\linewidth} 
\includegraphics[trim=0 0 0 0,scale=0.5,width=\columnwidth]{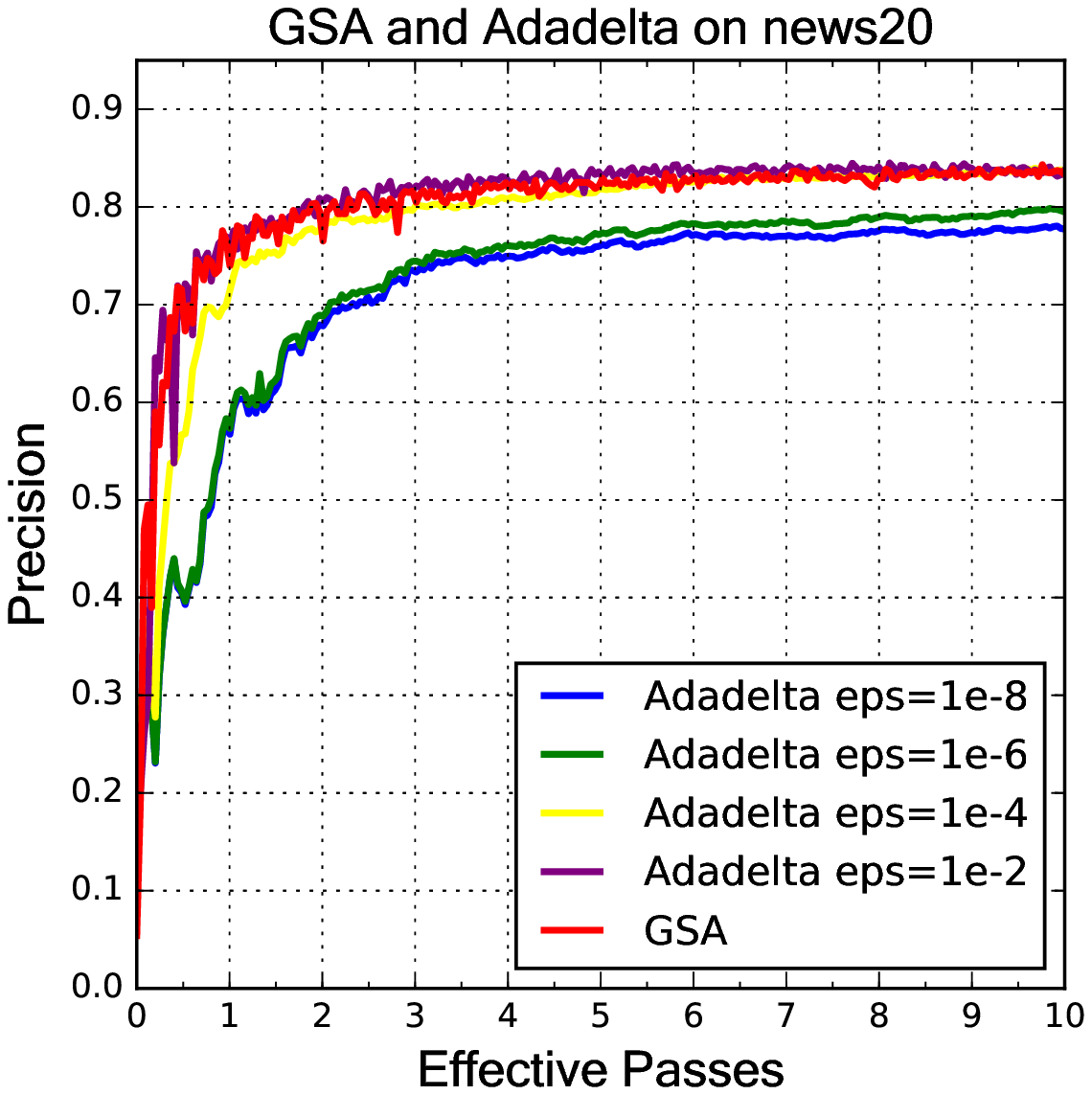} 
\end{minipage} 

\begin{minipage}[t]{0.41\linewidth} 
\includegraphics[trim=0 0 0 0,scale=0.5,width=\columnwidth]{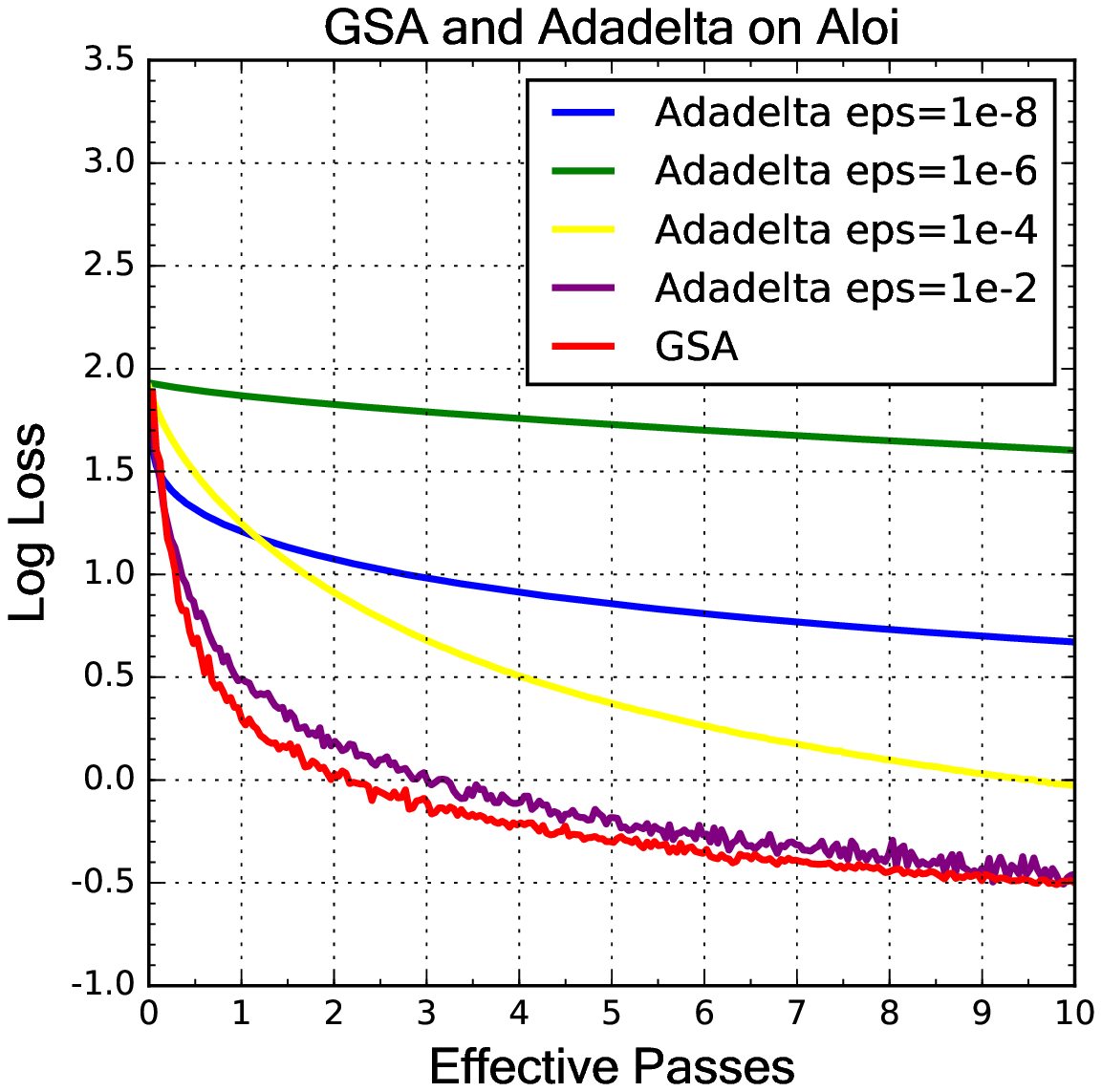} 
\end{minipage} 
\begin{minipage}[t]{0.41\linewidth} 
\includegraphics[trim=0 0 0 0,scale=0.5,width=\columnwidth]{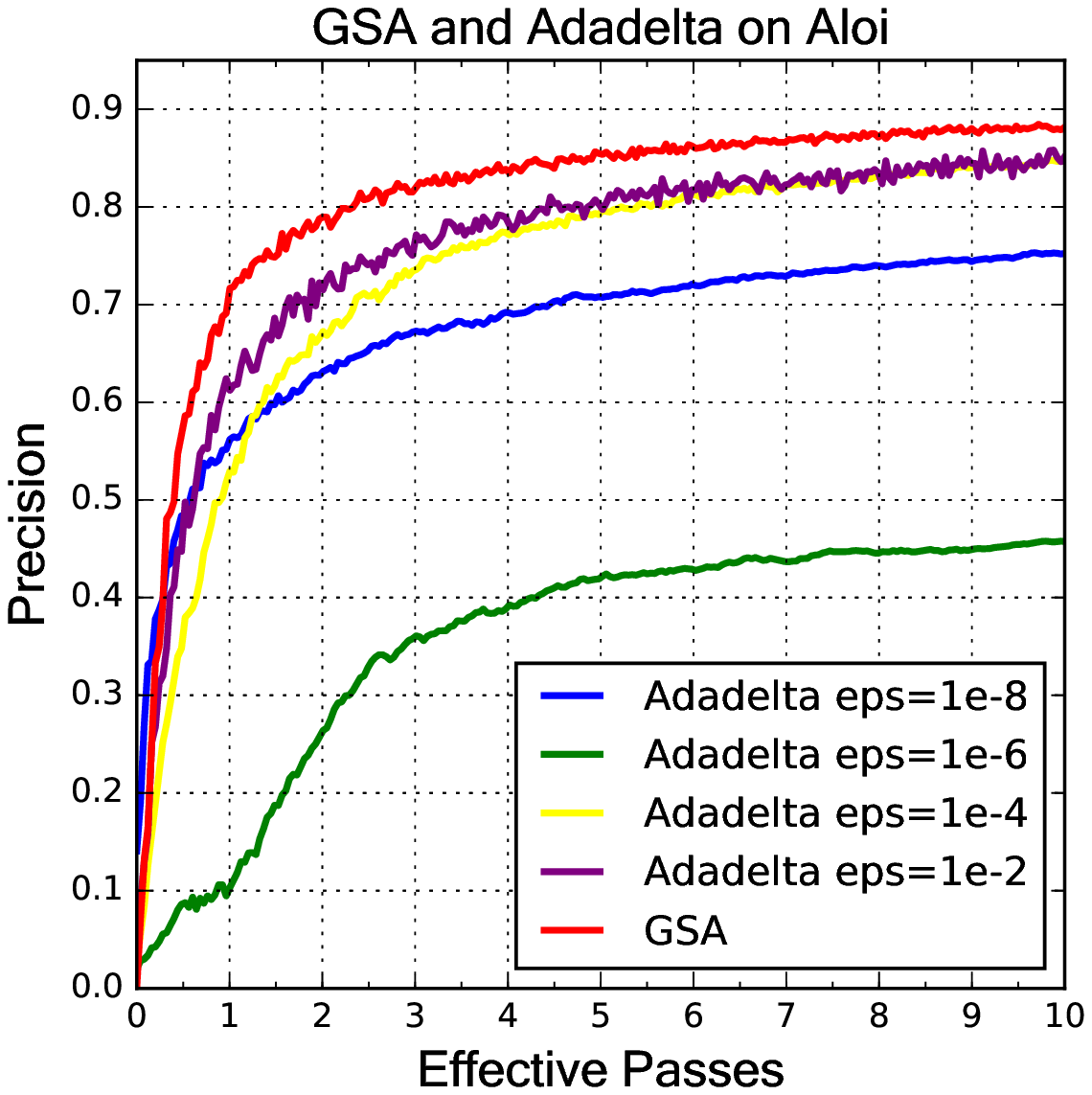} 
\end{minipage} 

\caption{\label{fig3}The performance of GSA and Adadelta}
\end{figure}

Adadelta has two hyper parameters: the decay rate $\gamma$ in (\ref{adadelta_gamma0}) and (\ref{adadelta_gamma}), the small constant $\varepsilon$ in (\ref{adadelta_eps}). Through our experiment we set $\gamma = (t-1)/t$ to replace the running window average of $\mathbb{E}[\Delta \theta^2]_t$ and $\mathbb{E}[g^2]_t$ to the whole time average. From Fig.\ref{fig3} we see that the performance of Adadelta significantly depends on $\varepsilon$, which is not like what the author claimed in \cite{adadelta}. In comparison, GSA is more robust on a variety of datasets. 
We also note that adadelta runs slower than GSA because its per-dimension learning rate introduce an extra $O(m)$ cost in time and space. On the other hand, GSA inherits the advantage of SGD in terms of computational efficiency. 

\subsection{Comparison with SCSG}
\begin{figure}[H]
\centering
\begin{minipage}[t]{0.41\linewidth} 
\includegraphics[trim=0 0 0 0,scale=0.5,width=\columnwidth]{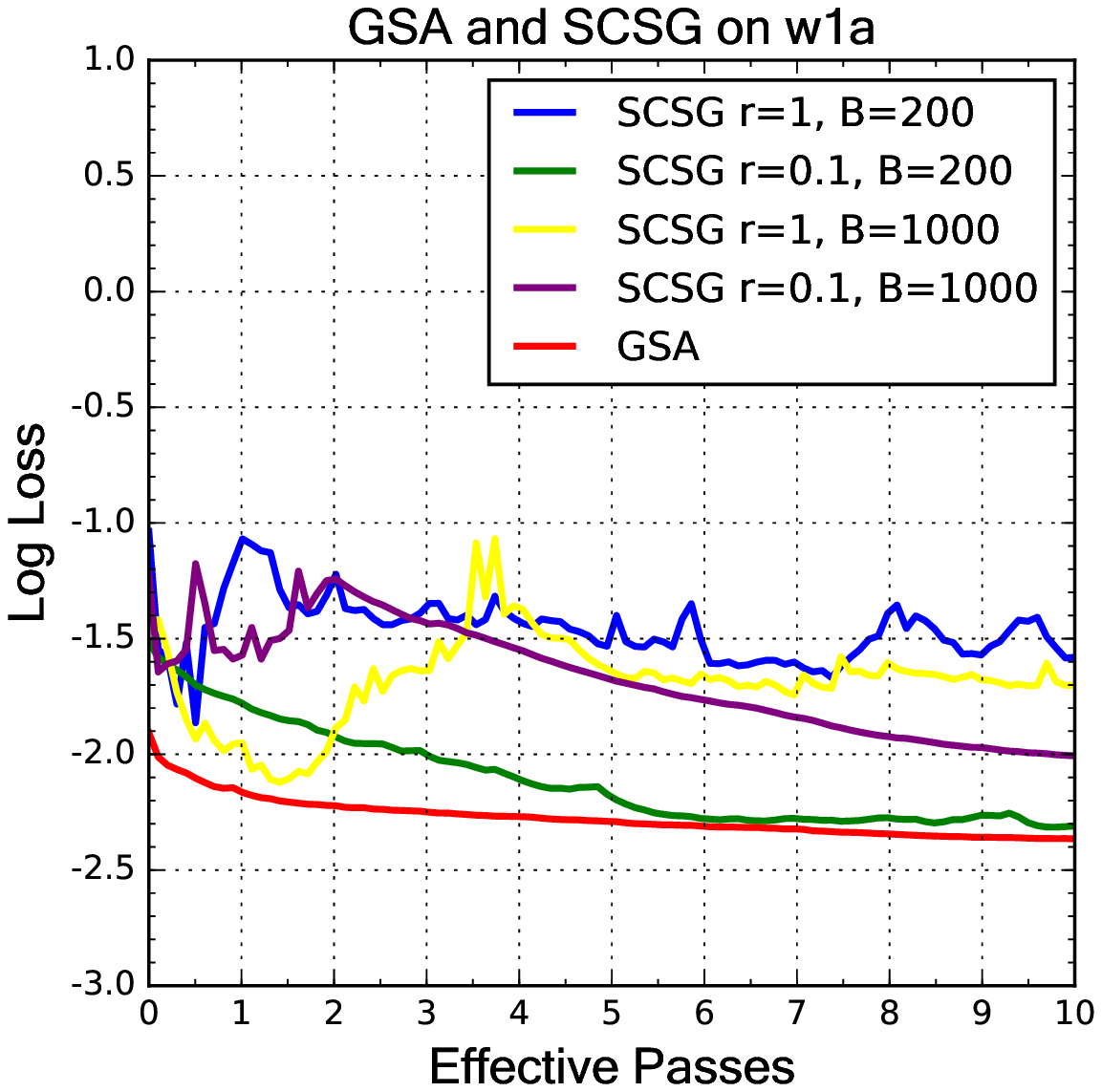} 
\end{minipage} 
\begin{minipage}[t]{0.41\linewidth} 
\includegraphics[trim=0 0 0 0,scale=0.5,width=\columnwidth]{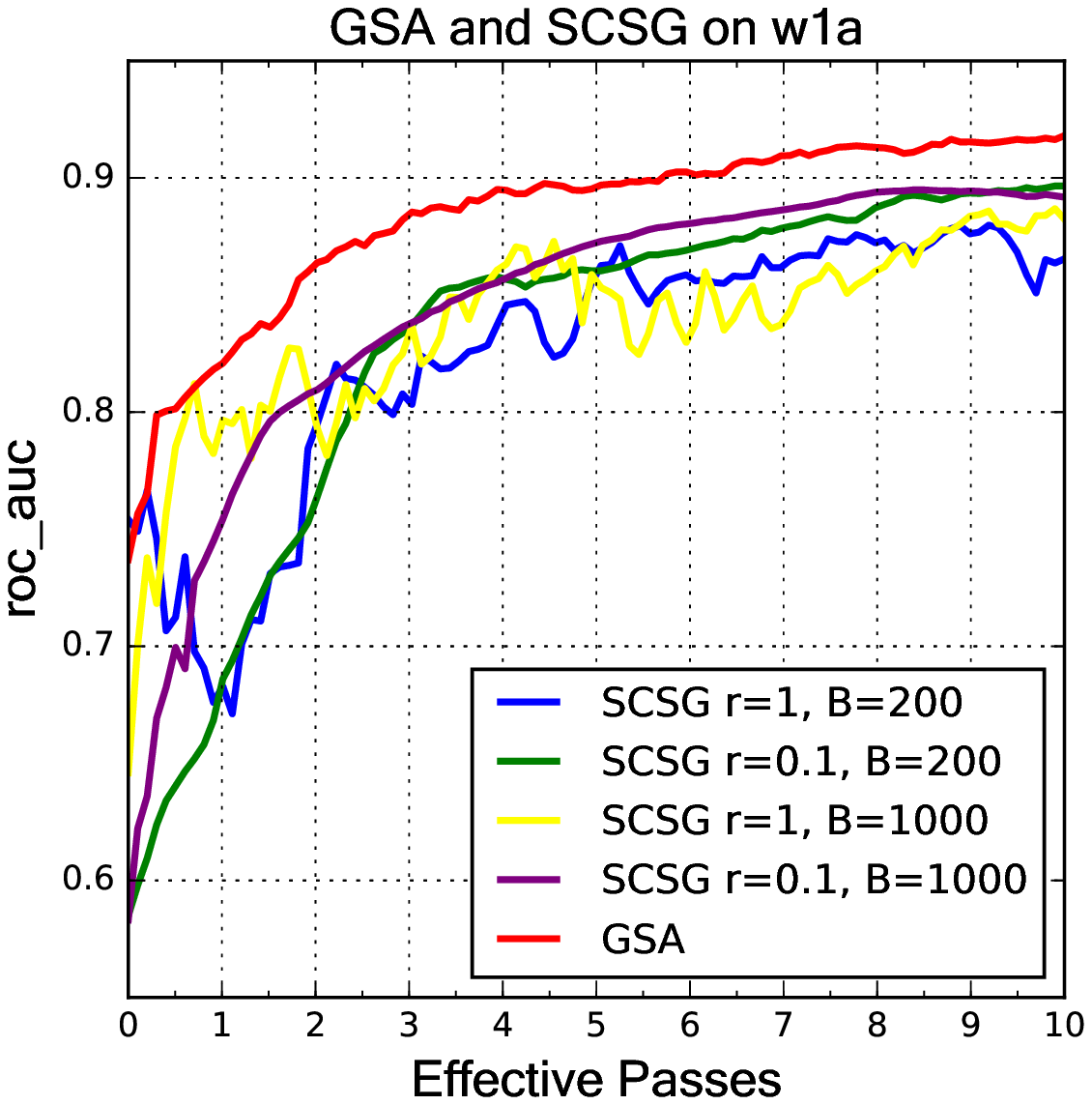} 
\end{minipage} 

\begin{minipage}[t]{0.41\linewidth} 
\includegraphics[trim=0 0 0 0,scale=0.5,width=\columnwidth]{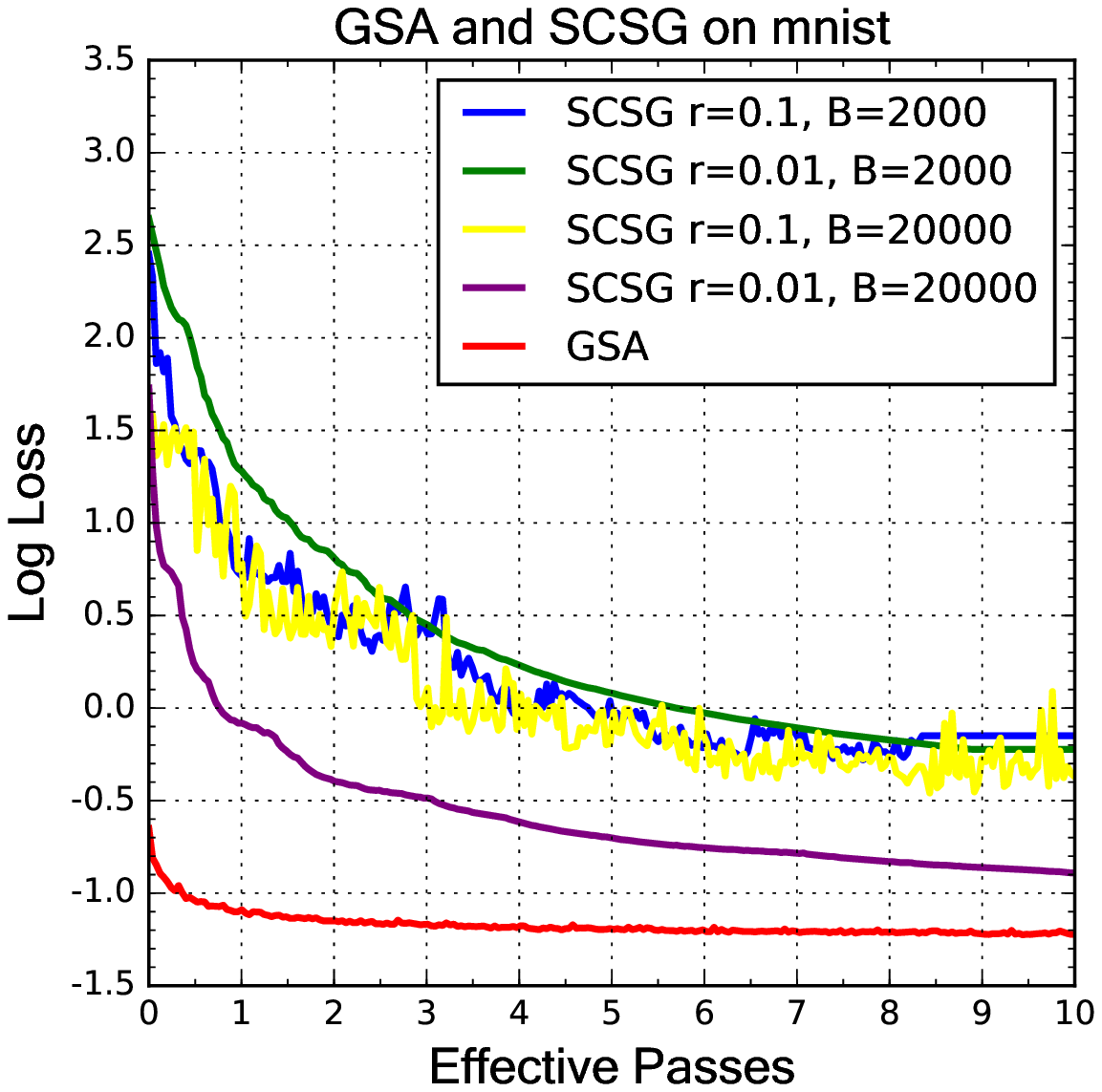} 
\end{minipage} 
\begin{minipage}[t]{0.41\linewidth} 
\includegraphics[trim=0 0 0 0,scale=0.5,width=\columnwidth]{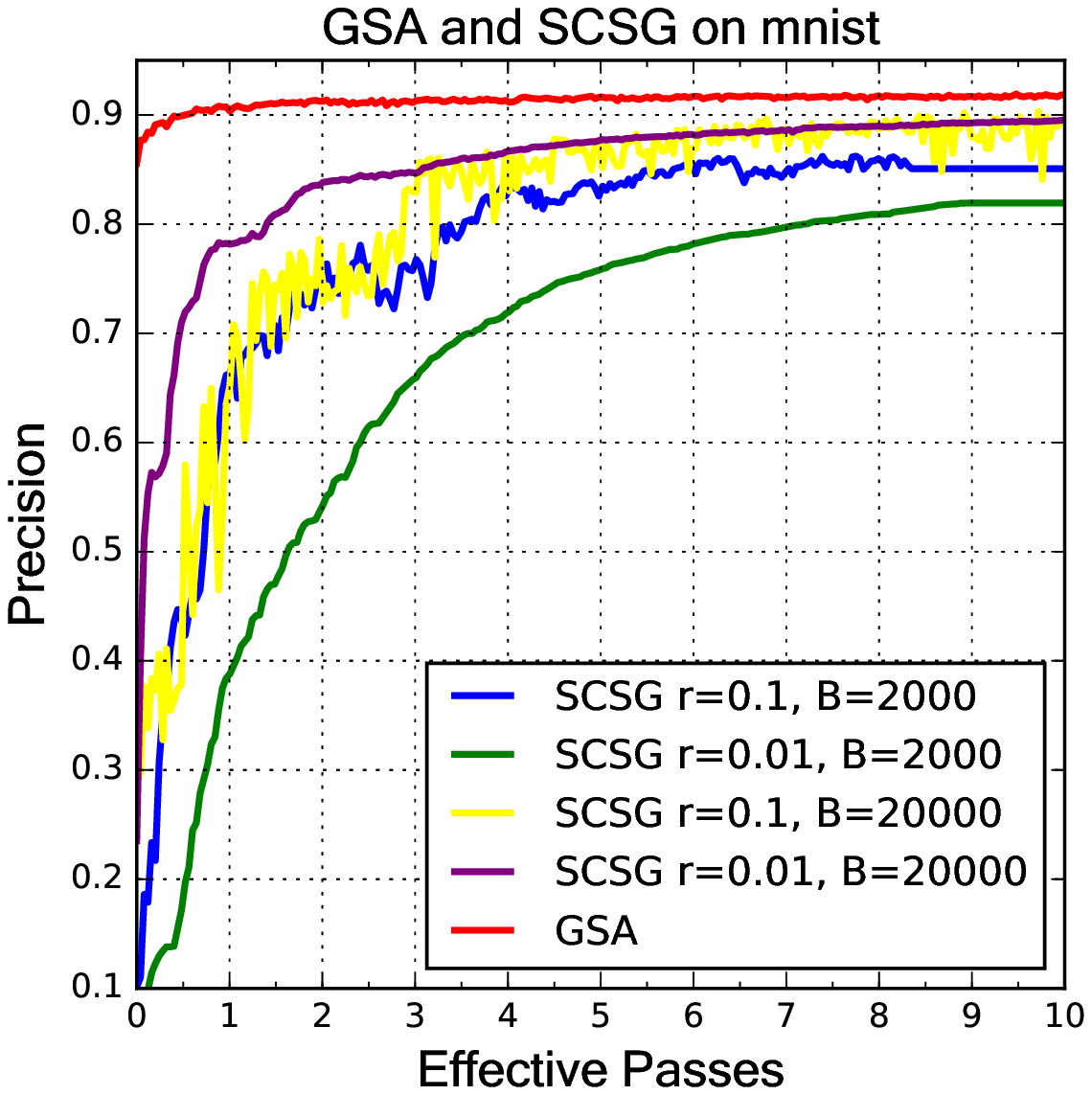} 
\end{minipage} 

\begin{minipage}[t]{0.41\linewidth} 
\includegraphics[trim=0 0 0 0,scale=0.5,width=\columnwidth]{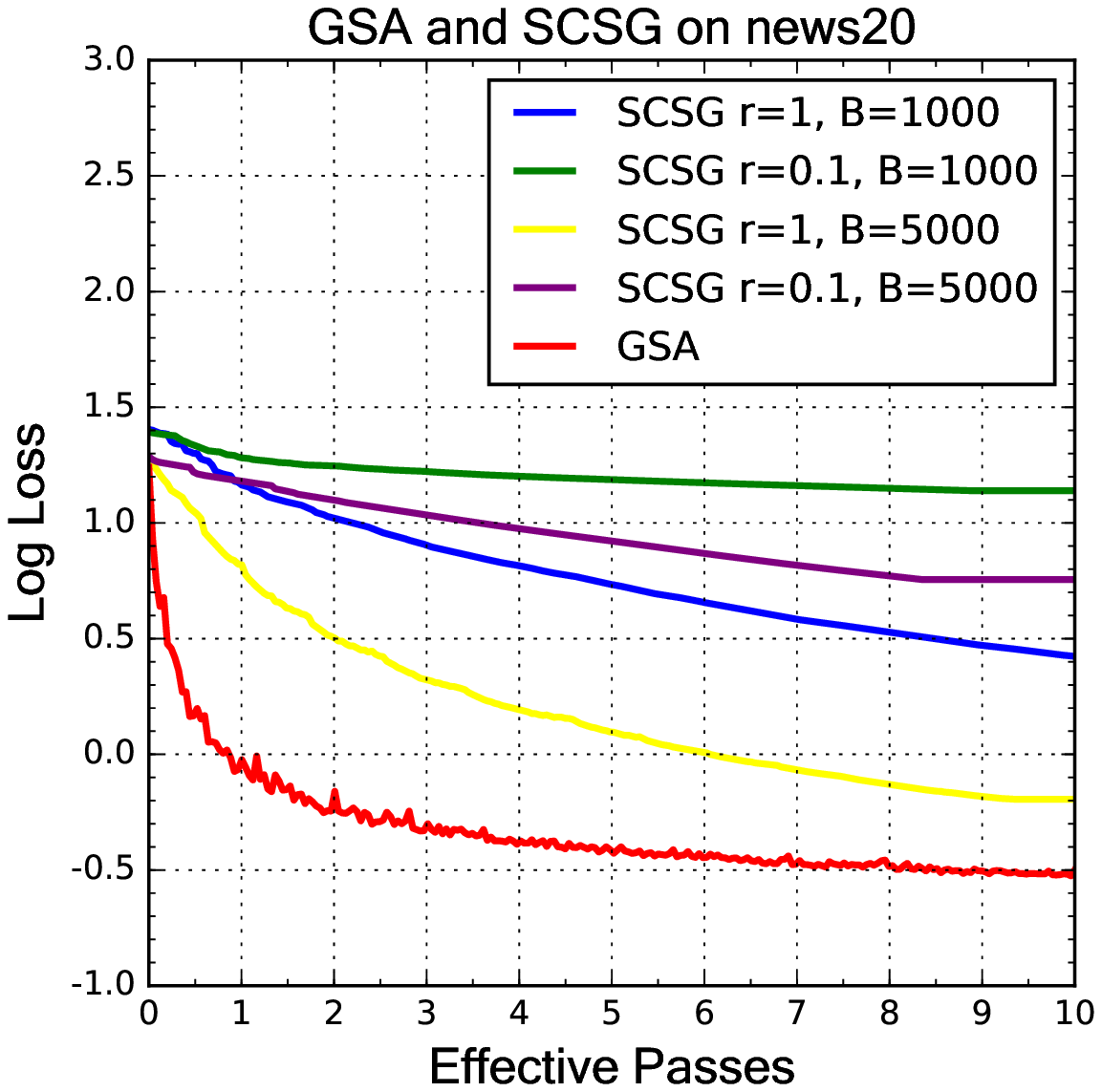} 
\end{minipage} 
\begin{minipage}[t]{0.41\linewidth} 
\includegraphics[trim=0 0 0 0,scale=0.5,width=\columnwidth]{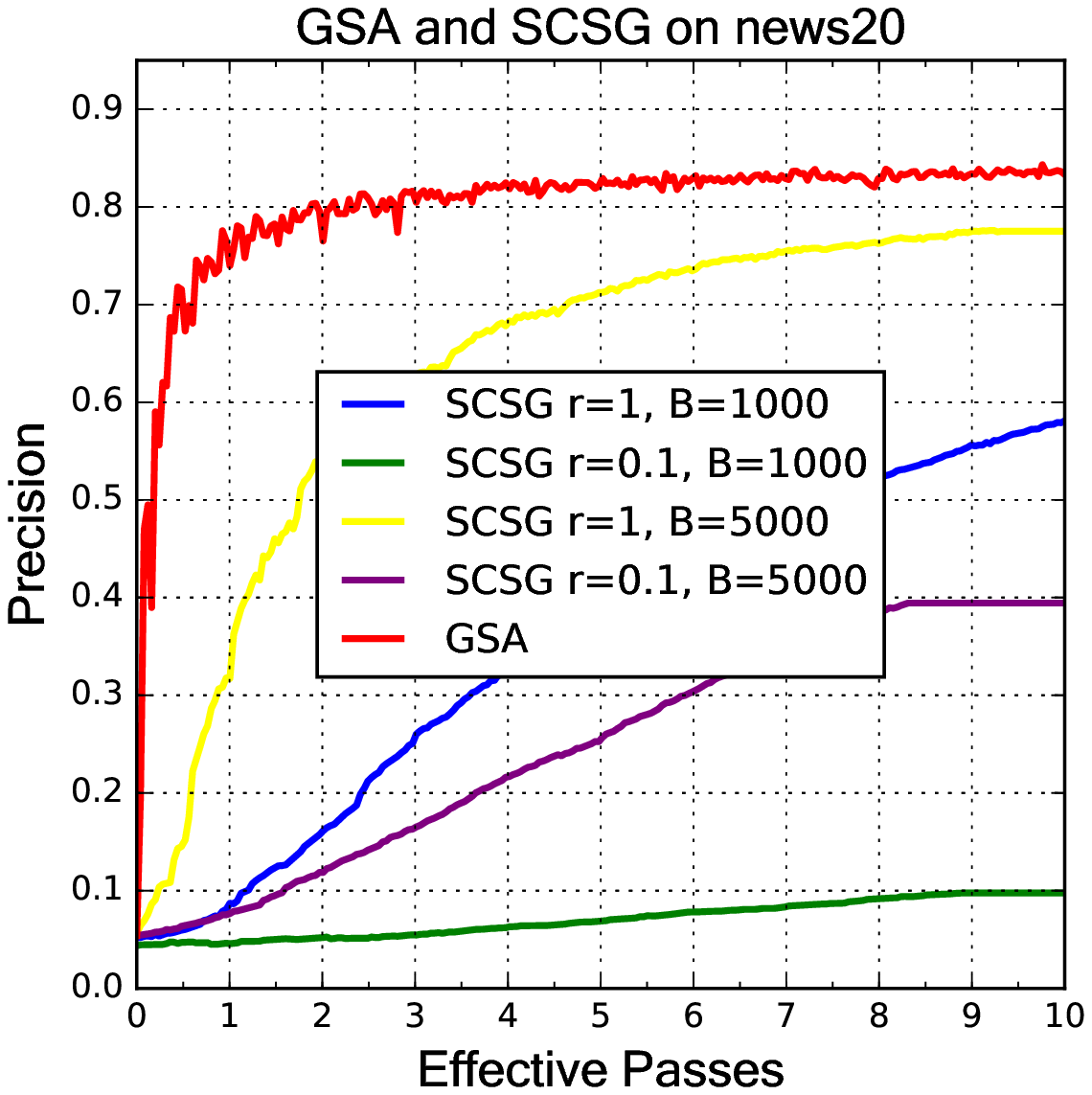} 
\end{minipage} 

\begin{minipage}[t]{0.41\linewidth} 
\includegraphics[trim=0 0 0 0,scale=0.5,width=\columnwidth]{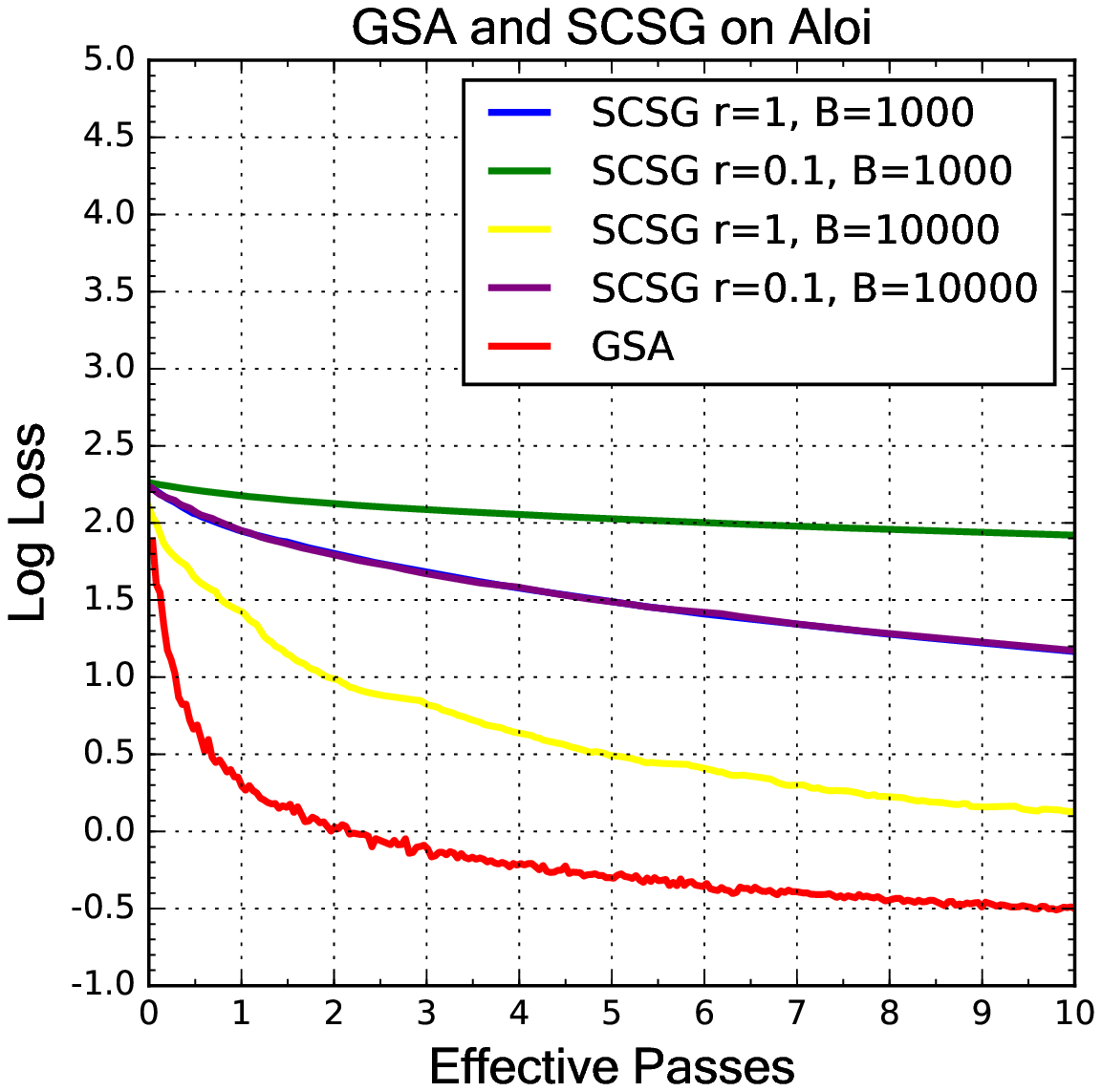} 
\end{minipage} 
\begin{minipage}[t]{0.41\linewidth} 
\includegraphics[trim=0 0 0 0,scale=0.5,width=\columnwidth]{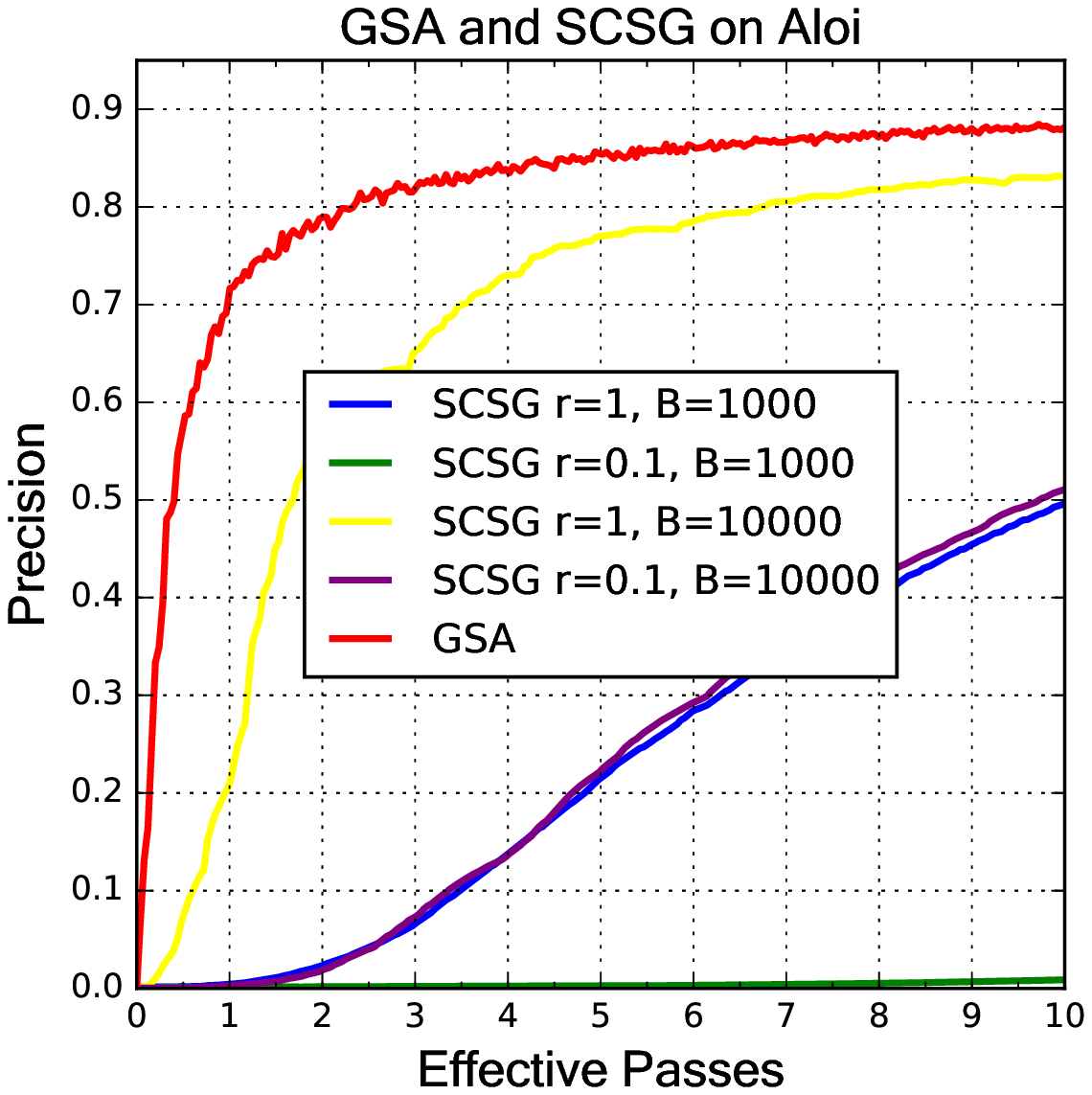} 
\end{minipage} 
\caption{\label{fig4}The performance of GSA and SCSG}
\end{figure}

SCSG is a variation of SVRG. Instead of the evaluation the full gradient, SCSG needs only a batch gradient during the beginning of each epoch. Therefore apart from learning rate $r$, the batch size $B$ also becomes an important hyper parameter. From Fig.\ref{fig4} we can see that better performance of SCSG is always associated with larger $B$, which undermines its applicability. On mnist and news20, the acceptable value of $B$ is approximately $1/3$ of the training size(20000/60000, 5000/15935). In addition, SCSG is also sensitive to step size $r$. Due to its sensitivity to the hyper parameters, the linear convergence rate of SCSG is hard to display because the optimal combination of parameter cannot be easily captured. In our experiment on the four datasets, GSA outperforms SCSG in each configuration.

\section{Discussion}
We propose GSA as a swift and parameter free stochastic optimization algorithm and test it on logistic and softmax regression. We show that GSA is facile in implementation and requires no extra memory. The experimental results on multiple datasets demonstrate that GSA is able to reach the performance of best tuned SGD and can also beat several state of the art algorithms in sense of practicability. Parameter tuning is cumbersome and rely on the priori on datasets to a large extent. If we do not have any domain knowledge of our data or problem, which is always the actual situation, the best way for us to optimize our algorithm is performing grid search. However, when confronted with large scaled data, it turns to be completely intractable. Therefore in industry, people tend to select model parameters based on practical experience. Unfortunately, most current advanced stochastic optimization methods are incapable of beating the best tuned basic SGD; they even crush from time to time. On the other hand, although GSA cannot always outperform the best performance of other prevalent methods, its convergence behaviour is stable and exhibits a comparable performance level of best tuned SGD. Thus we claim that GSA is more practical in application. 

In addition, we propose an easy implementation for parallelizing GSA. It is not an novel idea, but it works particularly well for GSA since it requires little I/O cost. On spark cluster our applications achieve a great success: For logistic regression model with $10^9$ instances and $10^7$ features, our method converges after only a single pass of dataset within 10 minutes(Source code on Github, see \cite{Fregata}). 

It should be noted that GSA is essentially basic SGD with an automatically selected learning rate sequence. However, unlike the two popular learning rate selection strategies, the constant step size rule and decreasing step size rule, the typical behaviour of GSA's step sizes displays a profile between them: it's basically a diminishing sequence asymptotically decrease towards a non-zero constant $a$. We believe $a$ is somehow around the best tuned SGD step-size, however the theoretical proof has not been established yet.

Since the step sizes of GSA do not necessarily converge to zero, there are several averaging scheme over iteration points to reduce the variance \cite{returnAvr, returnRunAvr, returnOptAvr}. Suppose that during each iteration, GSA returns a sequence of points $\omega_1, . . . , \omega_T$. To obtain a final estimation of $\omega^*$, a simple strategy is to return the last point $\omega_T$. It is well-acknowledged that this procedure has a $O(1/T)$ convergence guarantee for SGD. Another procedure, for which the standard online analysis of SGD applies \cite{returnAvr}, is to return the average point
$$
\bar{\omega}_T = \frac{1}{T}(\omega_1 + \cdots + \omega_T).
$$
However, the error bound for this procedure is merely $O(log(T)/T)$. Recently, in \cite{returnOptAvr} they indicate that  $O(log(T)/T)$ is not the best that one can achieve for strongly convex stochastic problems and have proposed a different algorithm, which is somewhat similar to SGD to achieve the $O(1/T)$ bound. Besides, in \cite{returnRunAvr} they present a new average tactics called $\alpha$-suffix averaging to recover the $O(1/T)$ rate. Since GSA is originated from the SGD framework, we recommend these averaging schemes as practicable options. However, it will incur an extra storage cost. For most cases without a specific accuracy target, we can just ignore this trick and return the last $\omega_T$. 

There are still several open issues remaining about our work. First, the current version of GSA only supports logistic and softmax regression. Actually, it can easily be generalized to other machine learning algorithm like SVM and linear regression, as long as the unique zero of its objective loss function has a closed form, at least in the sense of approximation. This is however not the case in neural networks. Because the loss function of a neural network is a black box: we only have access to the its value and gradient, which becomes an insurmountable obstacle to apply GSA. Even though a compromised strategy is to conduct linear search for a approximated greedy step size, the cost seems formidable since we cannot afford too many function evaluations within a single iteration step. We plan to explore several variants of GSA in future work to deal with this difficulty.

\noindent $\bm{Acknowledgements}$ We wish to thank Yafang Luo and Liran Chen for polishing vocabularies and sentences in this paper, and Yao Lu for several helpful discussions regarding this work.

\bibliographystyle{alpha}
\bibliography{sample}

\newpage 
\section{Appendix: More experimental results}
\begin{table}[H]
 \centering\small
 \caption{test on w1a}
\begin{tabular}{|c|c|c|c|c|c|c|c|c|c|c|}
\hline
 \multirow{2}{*}{Algo.} &  \multirow{2}{*}{hyper-para.} & \multicolumn{3}{|c|}{loss} & \multicolumn{3}{|c|}{prec.} & \multicolumn{3}{|c|}{auc}\\ \cline{3-11}
&  & 1 & 2 & 5 & 1 & 2 & 5 & 1 & 2 & 5\\ \hline
 \multicolumn{2}{|c|}{$\bm{best}$} & 0.086 & 0.086 & 0.079 & 0.977 & 0.979 & 0.979 & 0.911 & 0.916 & 0.918 \\ \hline
GSA & / & ~0.109~ & ~0.103~ & ~0.093~ & ~0.971~ & ~0.971~ & ~0.972~ & ~0.858~ & ~0.891~ & ~$\bm{0.918}$~\\ \hline
\multirow{5}{*}{SGD}  
& r=5 & 0.373 & 0.387 & 0.376 & 0.965 & 0.977 & 0.965 & 0.806 & 0.792 & 0.813 \\
& r=1 &  0.102 & 0.135 & 0.124 & $\bm{0.977}$ & $\bm{0.979}$ & 0.977 & 0.878 & 0.893 & 0.876\\
& r=.1 & 0.092 & $\bm{0.086}$ & $\bm{0.079}$ & 0.973 & 0.975 & 0.978 & $\bm{0.911}$ & $\bm{0.916}$ & 0.916\\
& r=.01 & 0.126 & 0.115 & 0.099 & 0.970 & 0.970 & 0.972  & 0.772 & 0.823 & 0.900 \\
& r=.001 & 0.271 & 0.196 & 0.144 & 0.970 & 0.970 & 0.970& 0.704 & 0.713 & 0.742\\
\hline
\multirow{5}{*}{Adadelta} 
& eps = 1e-8  & 0.693 & 0.692 & 0.691 & 0.970 & 0.970 & 0.970 & 0.698 & 0.697 & 0.698\\
& eps = 1e-6  & 0.659 & 0.622 & 0.528 & 0.970 & 0.970 & 0.970 & 0.699 & 0.698 & 0.703\\
& eps = 1e-4 & 0.189 & 0.144 & 0.111 & 0.970 & 0.970 & 0.971 & 0.735 & 0.767 & 0.853\\
& eps = 1e-2 & $\bm{0.086}$ & 0.092 & 0.094 & 0.976 & 0.978 & $\bm{0.979}$  & 0.892 & 0.906 & 0.890 \\
& eps = 1e-1  &  0.100 & 0.129 & 0.126 & $\bm{0.977}$ & $\bm{0.979}$ & 0.973 & 0.874 & 0.894 & 0.873\\
\hline
\multirow{4}{*}{SCSG} 
& r=1, B=200 & 0.150 & 0.180 & 0.200 & 0.972 & 0.975 & 0.972 & 0.832 & 0.835 & 0.857\\
& r=.1, B=200 & 0.124 & 0.109 & 0.091 & 0.970 & 0.973 & 0.975& 0.751 & 0.840 & 0.872 \\
& r=1, B=1000&0.593 & 0.439 & 0.414 & 0.962 & 0.964 & 0.958 & 0.821 & 0.861 & 0.859 \\
& r=.1, B=1000 &0.163 & 0.116 & 0.092 & 0.963 & 0.971 & 0.976& 0.766 & 0.846 & 0.883\\
\hline

\end{tabular}
\end{table}

\begin{table}[H]
 \centering\small
 \caption{test on mnist.scale}
\begin{tabular}{|c|c|c|c|c|c|c|c|}
\hline
 \multirow{2}{*}{Algo.} &  \multirow{2}{*}{hyper-para.} & \multicolumn{3}{|c|}{loss} & \multicolumn{3}{|c|}{prec.} \\ \cline{3-8}
&  & 1 & 2 & 5 & 1 & 2 & 5 \\ \hline
\multicolumn{2}{|c|}{$\bm{best}$} & 0.318 & 0.310 & 0.306 & 0.912 & 0.914 & 0.914 \\ \hline
GSA & / & ~~0.334~~ & ~~0.320~~ & ~~$\bm{0.306}$~~ & ~~0.905~~ & ~~0.907~~ & ~~$\bm{0.914}$~~ \\ \hline
\multirow{4}{*}{SGD}  
& r=.5 & 3.715 & 3.398 & 2.791 & 0.843 & 0.840 & 0.880  \\
& r=.1 & 1.347 & 1.131 & 0.995 & 0.819 & 0.840 & 0.876 \\
& r=.01 & 0.332 & 0.322 & 0.318 & 0.906 & 0.907 & 0.912\\
& r=.001 & 0.408 & 0.362 & 0.324 & 0.888 & 0.898 & 0.909 \\
\hline
\multirow{4}{*}{Adadelta} 
& eps = 1e-8  & 2.258 & 2.209 & 2.068 & 0.672 & 0.693 & 0.731 \\
& eps = 1e-6  & 0.889 & 0.647 & 0.466 & 0.835 & 0.859 & 0.882 \\
& eps = 1e-4 &  $\bm{0.318}$ & $\bm{0.310}$ & 0.319 & $\bm{0.912}$ & $\bm{0.914}$ & 0.913\\
& eps = 1e-2 & 4.507 & 3.133 & 2.618 & 0.792 & 0.857 & 0.881 \\
\hline
\multirow{4}{*}{SCSG} 
& r=.1, B=2000 & 4.254 & 2.797 & 1.336 & 0.443 & 0.615 & 0.794\\
& r=.01, B=2000 & 3.841 & 2.109 & 1.124 & 0.356 & 0.578 & 0.755\\
& r=.1, B=20000 & 4.932 & 2.343 & 0.773 & 0.339 & 0.548 & 0.872\\
& r=.01, B=20000 & 0.945 & 0.607 & 0.467 & 0.770 & 0.854 & 0.885\\
\hline
\end{tabular}
\end{table}

\begin{table}
 \centering\small
 \caption{test on news20.scale}
\begin{tabular}{|c|c|c|c|c|c|c|c|}
\hline
 \multirow{2}{*}{Algo.} &  \multirow{2}{*}{hyper-para.} & \multicolumn{3}{|c|}{loss} & \multicolumn{3}{|c|}{prec.} \\ \cline{3-8}
&  & 1 & 2 & 5 & 1 & 2 & 5 \\ \hline
\multicolumn{2}{|c|}{$\bm{best}$} & 0.973 & 0.787 & 0.616 & 0.761 & 0.806 & 0.830 \\ \hline
GSA & / & ~~0.977~~ & ~~0.852~~ & ~~0.652~~ & ~~0.740~~ & ~~0.765~~ & ~~$\bm{0.830}$~~ \\ \hline
\multirow{5}{*}{SGD}  
& r=5 & $\bm{0.973}$ & 1.213 & 0.817 & 0.740 & 0.747 & 0.812 \\
& r=1 &   1.165 & 0.842 & 0.617 & 0.658 & 0.777 & 0.825\\
& r=.1 &2.397 & 1.956 & 1.321 & 0.532 & 0.717 & 0.772 \\
& r=.01 & 2.927 & 2.858 & 2.671 & 0.104 & 0.607 & 0.671\\
& r=.001 & 2.989 & 2.982 & 2.961 & 0.051 & 0.054 & 0.401 \\
\hline
\multirow{5}{*}{Adadelta} 
& eps = 1e-8  &  2.996 & 2.996 & 2.995 & 0.567 & 0.678 & 0.761\\
& eps = 1e-6  &  2.987 & 2.977 & 2.941 & 0.573 & 0.688 & 0.773\\
& eps = 1e-4 & 2.432 & 1.965 & 1.230 & 0.715 & 0.786 & 0.818 \\
& eps = 1e-2 & 1.072 & 0.801 & $\bm{0.616}$ & $\bm{0.761}$ & $\bm{0.806}$ & $\bm{0.830}$ \\
& eps = 1e-1 & 1.047 & $\bm{0.787}$ & 0.632 & 0.747 & 0.799 & 0.819 \\
\hline
\multirow{4}{*}{SCSG} 
& r=1, B=1000 & 3.202 & 2.773 & 2.081 & 0.087 & 0.161 & 0.392\\
& r=.1, B=1000 & 3.684 & 3.635 & 3.527 & 0.046 & 0.047 & 0.054\\
& r=1, B=5000 & 2.268 & 1.652 & 1.100 & 0.317 & 0.538 & 0.712\\
& r=.1, B=5000 & 3.558 & 3.482 & 3.318 & 0.057 & 0.058 & 0.064 \\
\hline
\end{tabular}
\end{table}

\begin{table}
 \centering\small
 \caption{test on aloi.scale}
\begin{tabular}{|c|c|c|c|c|c|c|c|}
\hline
 \multirow{2}{*}{Algo.} &  \multirow{2}{*}{hyper-para.} & \multicolumn{3}{|c|}{loss} & \multicolumn{3}{|c|}{prec.} \\ \cline{3-8}
&  & 1 & 2 & 5 & 1 & 2 & 5 \\ \hline
\multicolumn{2}{|c|}{$\bm{best}$} & 1.029 & 0.812 & 0.606 & 0.790 & 0.835 & 0.882 \\ \hline
GSA & / & ~~$\bm{1.029}$~~ & ~~$\bm{0.812}$~~ & ~~$\bm{0.606}$~~ & ~~$\bm{0.790}$~~ & ~~$\bm{0.835}$~~ & ~~$\bm{0.882}$~~ \\ \hline
\multirow{4}{*}{SGD}  
& r=3 &  2.144 & 2.130 & 1.577 & 0.701 & 0.730 & 0.787\\
& r=1 & 1.196 & 0.938 & 0.659 & 0.727 & 0.803 & 0.851  \\
& r=.1 &2.485 & 1.740 & 1.113 & 0.631 & 0.736 & 0.821 \\
& r=.01 &5.558 & 4.695 & 3.340 & 0.289 & 0.454 & 0.656 \\
\hline
\multirow{4}{*}{Adadelta} 
& eps = 1e-8  & 2.926 & 2.488 & 1.957 & 0.632 & 0.691 & 0.752 \\
& eps = 1e-6  & 6.207 & 5.797 & 4.967 & 0.264 & 0.393 & 0.458 \\
& eps = 1e-4 & 2.483 & 1.656 & 0.974 & 0.673 & 0.771 & 0.846 \\
& eps = 1e-2 &  1.207 & 0.898 & 0.633 & 0.715 & 0.782 & 0.851\\
\hline
\multirow{4}{*}{SCSG} 
& r=1, B=1000 & 6.051 & 4.838 & 3.210 & 0.024 & 0.139 & 0.496\\
& r=.1, B=1000 &8.364 & 7.804 & 6.824 & 0.002 & 0.002 & 0.009 \\
& r=1, B=10000 & 2.684 & 1.887 & 1.135 & 0.566 & 0.731 & 0.831\\
& r=.1, B=10000 & 6.000 & 4.855 & 3.230 & 0.019 & 0.138 & 0.511 \\
\hline
\end{tabular}
\end{table}

\begin{table}
 \centering\small
 \caption{test on a9a}
\begin{tabular}{|c|c|c|c|c|c|c|c|c|c|c|}
\hline
 \multirow{2}{*}{Algo.} &  \multirow{2}{*}{hyper-para.} & \multicolumn{3}{|c|}{loss} & \multicolumn{3}{|c|}{prec.} & \multicolumn{3}{|c|}{auc}\\ \cline{3-11}
&  & 1 & 2 & 5 & 1 & 2 & 5 & 1 & 2 & 5\\ \hline
\multicolumn{2}{|c|}{$\bm{best}$} & 0.328  & 0.326 & 0.325 & 0.847 & 0.848  & 0.849 & 0.902 & 0.904 & 0.904 \\ \hline
GSA & / & ~0.330~ & ~0.328~ & ~0.327~ & ~0.846~ & ~ $\bm{0.848}$~ & ~0.847~ & ~ $\bm{0.902}$~ & ~0.903~ & ~0.903~\\ \hline
\multirow{5}{*}{SGD}  
& r=5 &  5.645 & 4.297 & 4.273 & 0.760 & 0.803 & 0.821 & 0.814 & 0.825 & 0.736\\
& r=1 &  1.927 & 1.292 & 1.015 & 0.733 & 0.792 & 0.825 & 0.858 & 0.872 & 0.867  \\
& r=.1 & 0.380 & 0.350 & 0.351 & 0.820 & 0.841 & 0.838 & 0.894 & 0.897 & 0.894 \\
& r=.01 & $\bm{0.328}$ & 0.327 & 0.326 & 0.846 & 0.847 & 0.848 &  $\bm{0.902}$ & 0.903 &  $\bm{0.904}$ \\
& r=.001 & 0.353 & 0.340 & 0.329 & 0.836 & 0.841 & 0.846 & 0.889 & 0.896 & 0.901 \\
\hline
\multirow{5}{*}{Adadelta} 
& eps = 1e-8  & 0.688 & 0.683 & 0.667 & 0.759 & 0.759 & 0.759 & 0.818 & 0.822 & 0.828 \\
& eps = 1e-6  & 0.515 & 0.469 & 0.405 & 0.759 & 0.760 & 0.812 & 0.862 & 0.874 & 0.882 \\
& eps = 1e-4 & 0.328 & $\bm{0.326}$ & $\bm{0.325}$ & 0.846 & 0.847 &  $\bm{0.849}$ &  $\bm{0.902}$ & $\bm{0.904}$ &  $\bm{0.904}$ \\
& eps = 1e-2 & 0.594 & 0.546 & 0.535 & 0.772 & 0.807 & 0.817 & 0.871 & 0.876 & 0.869 \\
& eps = 1e-1  &  1.707 & 1.449 & 0.969 & 0.742 & 0.770 & 0.824 & 0.860 & 0.869 & 0.869 \\
\hline
\multirow{6}{*}{SCSG} 
& r=1, B=1000 & 1.217 & 1.115 & 1.101 & 0.791 & 0.819 & 0.800 & 0.843 & 0.849 & 0.848 \\
& r=.1, B=1000 & 0.338 & 0.346 & 0.344 & 0.842 & 0.844 & 0.839 & 0.897 & 0.896 & 0.897 \\
& r=.01, B=1000 &0.336 & 0.330 & 0.334 & 0.843 & 0.846 & 0.845 & 0.897 & 0.901 & 0.899 \\
& r=1, B=10000 &  0.878 & 0.918 & 1.099 & 0.820 & 0.826 & 0.828 & 0.868 & 0.875 & 0.863\\
& r=.1, B=10000 & 0.329 & 0.327 & 0.334 &  $\bm{0.847}$ &  0.847 & 0.845 &  $\bm{0.902}$ & 0.902 & 0.898\\
& r=.01, B=10000 & 0.337 & 0.330 & 0.327 & 0.842 & 0.846 & 0.848 & 0.896 & 0.901 & 0.903\\
\hline

\end{tabular}
\end{table}

\begin{table}
 \centering\small
 \caption{test on breast-cancer\_scale}
\begin{tabular}{|c|c|c|c|c|c|c|c|c|c|c|}
\hline
 \multirow{2}{*}{Algo.} &  \multirow{2}{*}{hyper-para.} & \multicolumn{3}{|c|}{loss} & \multicolumn{3}{|c|}{prec.} & \multicolumn{3}{|c|}{auc}\\ \cline{3-11}
&  & 1 & 2 & 5 & 1 & 2 & 5 & 1 & 2 & 5\\ \hline
\multicolumn{2}{|c|}{$\bm{best}$} & 0.088 & 0.085 & 0.079 & 0.974 & 0.975 & 0.974 & 0.996 & 0.996 & 0.996\\ \hline
 GSA & / & ~0.109~ & ~0.100~ & ~0.090~ & ~0.959~ & ~0.962~ & ~0.968~ & ~ $\bm{0.996}$~ & ~$\bm{0.996}$~ & ~ $\bm{0.996}$~\\ \hline
\multirow{4}{*}{SGD}  
& r=5 & 0.401 & 1.302 & 0.520 &  $\bm{0.974}$ & 0.814 & 0.946 & 0.992 & 0.982 & 0.988  \\
& r=1 & 0.099 & 0.307 & 0.180 &  $\bm{0.974}$ & 0.903 & 0.949 & 0.996 & 0.993 & 0.994  \\
& r=.1 &  $\bm{0.088}$ &  $\bm{0.085}$ & 0.083 & 0.971 & 0.971 &  $\bm{0.974}$ & 0.996 & 0.996 & 0.996 \\
& r=.01  & 0.171 & 0.131 & 0.100 & 0.944 & 0.958 & 0.965 & 0.996 & 0.996 & 0.996\\
\hline
\multirow{4}{*}{Adadelta} 
& eps = 1e-6  & 0.682 & 0.670 & 0.632 & 0.843 & 0.845 & 0.843 & 0.995 & 0.995 & 0.995\\
& eps = 1e-4 & 0.331 & 0.250 & 0.163 & 0.884 & 0.921 & 0.944 & 0.995 & 0.996 & 0.996 \\
& eps = 1e-2 & 0.089 & 0.088 & $\bm{0.079}$ & 0.958 & 0.965 & 0.971 & 0.996 & 0.996 & 0.996 \\
& eps = 1e-1  & 0.112 & 0.106 & 0.094 & 0.958 & 0.959 & 0.971 & 0.996 & 0.995 & 0.996\\
\hline
\multirow{4}{*}{SCSG} 
& r=1, B=20 & 0.111 & 0.133 & 0.139 & 0.966 &  $\bm{0.975}$ & 0.971 & 0.996 & 0.996 & 0.995 \\
& r=.1, B=20  & 0.104 & 0.087 & 0.083 & 0.959 & 0.966 &  $\bm{0.974}$ & 0.995 & 0.996 & 0.996\\
& r=1, B=200 & 0.191 & 0.232 & 0.169 & 0.964 & 0.964 & 0.964 & 0.993 & 0.992 & 0.992 \\
& r=.1, B=200  & 0.096 & 0.088 & 0.081 & 0.959 & 0.962 & 0.972 & 0.995 & 0.995 & 0.996\\
\hline
\end{tabular}
\end{table}

\begin{table}
 \centering\small
 \caption{test on gistte\_scale}
\begin{tabular}{|c|c|c|c|c|c|c|c|c|c|c|}
\hline
 \multirow{2}{*}{Algo.} &  \multirow{2}{*}{hyper-para.} & \multicolumn{3}{|c|}{loss} & \multicolumn{3}{|c|}{prec.} & \multicolumn{3}{|c|}{auc}\\ \cline{3-11}
&  & 1 & 2 & 5 & 1 & 2 & 5 & 1 & 2 & 5\\ \hline
\multicolumn{2}{|c|}{$\bm{best}$} &0.165  & 0.162 & 0.105 & 0.962 & 0.966 & 0.972 & 0.991 & 0.993 & 0.993\\ \hline
GSA & / & ~0.229~ & ~0.192~ & ~0.169~ & ~0.938~ & ~0.944~ & ~0.944~ & ~0.983~ & ~0.986~ & ~0.989~\\ \hline
\multirow{4}{*}{SGD}  
& r=10 &  1.813 & 1.785 & 2.159 & 0.948 & 0.948 & 0.938 & 0.945 & 0.945 & 0.941\\
& r=1 &  1.612 & 1.625 & 2.779 & 0.953 & 0.952 & 0.917 & 0.951 & 0.951 & 0.924 \\
& r=.1 & 2.306 & 1.091 & 1.303 & 0.919 & 0.965 & 0.957 & 0.938 & 0.966 & 0.960 \\
& r=.01 & 0.391 & 0.356 & 0.424 &  $\bm{0.962}$ &  $\bm{0.966}$ & 0.962 &  0.987 & 0.988 & 0.988\\
\hline
\multirow{5}{*}{Adadelta} 
& eps = 1e-8 & 0.687 & 0.680 & 0.658 & 0.872 & 0.885 & 0.868 & 0.948 & 0.948 & 0.949\\
& eps = 1e-6  & 0.406 & 0.311 & 0.231 & 0.911 & 0.921 & 0.931 & 0.966 & 0.974 & 0.983 \\
& eps = 1e-4 & $\bm{0.165}$ & $\bm{0.162}$ &  $\bm{0.105}$ & 0.934 & 0.953 &  $\bm{0.972}$ & $\bm{0.991}$ &  $\bm{0.993}$ &  $\bm{0.993}$ \\
& eps = 1e-2 & 2.596 & 1.109 & 1.034 & 0.917 & 0.963 & 0.968 & 0.919 & 0.967 & 0.970 \\
& eps = 1e-1  & 2.504 & 2.647 & 1.048 & 0.925 & 0.922 & 0.969 & 0.932 & 0.918 & 0.971 \\
\hline
\multirow{6}{*}{SCSG} 
& r=1, B=200 & 3.071 & 1.308 & 3.313 & 0.909 & 0.962 & 0.902 & 0.905 & 0.962 & 0.910 \\
& r=.1, B=200 & 4.225 & 1.844 & 1.445 & 0.867 & 0.941 & 0.954 & 0.885 & 0.948 & 0.958\\
& r=.01, B=200 & 3.467 & 1.442 & 1.325 & 0.838 & 0.929 & 0.938 & 0.900 & 0.958 & 0.962\\
& r=1, B=2000 &  3.627 & 2.015 & 1.813 & 0.895 & 0.942 & 0.948 & 0.899 & 0.943 & 0.944\\
& r=.1, B=2000 & 3.378 & 1.412 & 2.835 & 0.895 & 0.952 & 0.911 & 0.907 & 0.955 & 0.923\\
& r=.01, B=2000 & 2.383 & 2.280 & 1.501 & 0.905 & 0.910 & 0.934 & 0.920 & 0.931 & 0.955 \\
\hline

\end{tabular}
\end{table}

\begin{table}
 \centering\small
 \caption{test on madelon}
\begin{tabular}{|c|c|c|c|c|c|c|c|c|c|c|}
\hline
 \multirow{2}{*}{Algo.} &  \multirow{2}{*}{hyper-para.} & \multicolumn{3}{|c|}{loss} & \multicolumn{3}{|c|}{prec.} & \multicolumn{3}{|c|}{auc}\\ \cline{3-11}
&  & 1 & 5 & 20 & 1 & 5 & 20 & 1 & 5 & 20\\ \hline
\multicolumn{2}{|c|}{$\bm{best}$} & 0.696 & 0.678 & 0.671 & 0.552 & 0.613 & 0.607 & 0.614 & 0.620 & 0.620 \\ \hline
GSA & / & ~0.952~ & ~0.690~ & ~$\bm{0.671}$~ & ~0.500~ & ~0.538~ & ~0.587~ & ~0.613~ & ~0.619~ & ~$\bm{0.620}$~\\ \hline
\multirow{5}{*}{SGD}  
& r=1e-4 &   17.269 & 14.927 & 16.242 & 0.500 & 0.565 & 0.527 & 0.500 & 0.566 & 0.527\\
& r=1e-5 &  17.269 & 13.162 & 13.008 & 0.500 & 0.605 & $\bm{0.607}$ & 0.500 & 0.599 & 0.607 \\
& r=1e-6 & 17.269 & 15.070 & 15.884 & 0.500 & 0.507 & 0.505 & 0.500 & 0.519 & 0.525 \\
& r=1e-7 & 3.099 & 2.322 & 2.219 & 0.500 & 0.507 & 0.523 & 0.611 & $\bm{0.620}$ & 0.618 \\
& r=1e-8 &1.011 & $\bm{0.678}$ & 0.672 & 0.500 & $\bm{0.613}$ & 0.590 & 0.613 & 0.619 & $\bm{0.620}$ \\
\hline
\multirow{5}{*}{Adadelta} 
& eps = 1e-8  & $\bm{0.696}$ & 0.705 & 0.689 & 0.500 & 0.500 & 0.507 & 0.608 & 0.615 & 0.619 \\
& eps = 1e-6  & 1.473 & 0.869 & 0.942 & 0.500 & 0.515 & 0.547 & $\bm{0.614}$ & $\bm{0.620}$ & 0.617 \\
& eps = 1e-4 &  17.269 & 16.805 & 16.193 & 0.500 & 0.507 & 0.515 & 0.500 & 0.512 & 0.523\\
& eps = 1e-2 &17.269 & 16.924 & 17.269 & 0.500 & 0.510 & 0.500 & 0.500 & 0.510 & 0.500 \\
& eps = 1e-1  &  17.269 & 16.809 & 15.024 & 0.500 & 0.513 & 0.565 & 0.500 & 0.513 & 0.565\\
\hline
\multirow{4}{*}{SCSG} 
& r=1e-6, B=200 & 18.421 & 18.213 & 17.336 & 0.450 & 0.462 & 0.480 & 0.453 & 0.460 & 0.478 \\
& r=1e-7, B=200 & 17.269 & 16.756 & 16.553 & 0.500 & 0.502 & 0.515 & 0.500 & 0.505 & 0.514\\
& r=1e-8, B=200 &  17.269 & 17.269 & 16.270 & 0.500 & 0.500 & 0.522 & 0.500 & 0.500 & 0.523\\
& r=1e-6, B=1000 & 15.221 & 15.395 & 15.663 & $\bm{0.552}$ & 0.543 & 0.532 & 0.548 & 0.542 & 0.536\\
& r=1e-7, B=1000 & 17.099 & 17.651 & 17.495 & 0.502 & 0.478 & 0.485 & 0.502 & 0.477 & 0.484 \\
& r=1e-8, B=1000 & 17.269 & 17.269 & 16.490 & 0.500 & 0.500 & 0.510 & 0.500 & 0.500 & 0.512\\
\hline

\end{tabular}
\end{table}

\begin{table}
 \centering\small
 \caption{test on cod-rna}
\begin{tabular}{|c|c|c|c|c|c|c|c|c|c|c|}
\hline
 \multirow{2}{*}{Algo.} &  \multirow{2}{*}{hyper-para.} & \multicolumn{3}{|c|}{loss} & \multicolumn{3}{|c|}{prec.} & \multicolumn{3}{|c|}{auc}\\ \cline{3-11}
&  & 1 & 2 & 5 & 1 & 2 & 5 & 1 & 2 & 5\\ \hline
\multicolumn{2}{|c|}{$\bm{best}$} & 0.637 & 0.623 & 0.620 & 0.890 & 0.890 & 0.899 & 0.931 & 0.932 & 0.933 \\ \hline
GSA & / & ~0.654~ & ~0.653~ & ~0.652~ & ~0.886~ & ~0.887~ & ~0.887~ & ~$\bm{0.931}$~ & ~$\bm{0.932}$~ & ~0.931~\\ \hline
\multirow{5}{*}{SGD}  
& r=5 & 0.738 & 0.715 & 0.739 & 0.716 & 0.751 & 0.714 & 0.800 & 0.821 & 0.799
 \\
& r=1 &  0.705 & 0.768 & 0.707 & 0.765 & 0.666 & 0.763 & 0.900 & 0.86 & 0.900
 \\
& r=.1 & 0.667 & 0.667 & 0.667 & 0.851 & 0.851 & 0.851 & 0.931 & 0.931 & 0.931
 \\
& r=.01 & 0.647 & 0.644 & 0.643 & 0.880 & 0.882 & 0.882 & 0.931 & 0.931 & 0.931
 \\
 & r=.001 & 0.701 & 0.678 & 0.658 & 0.861 & 0.879 & 0.885 & 0.931 & 0.931 & 0.931 
 \\
\hline
\multirow{5}{*}{Adadelta} 
& eps = 1e-8  &  0.643 & 0.639 & 0.638 & $\bm{0.890}$ & $\bm{0.890}$ & $\bm{0.889}$ & 0.931 & 0.931 & 0.931
\\
& eps = 1e-6  & 0.657 & 0.655 & 0.635 & 0.875 & 0.877 & 0.886 & 0.931 & 0.931 & 0.932

 \\
& eps = 1e-4 & $\bm{0.637}$ & 0.651 & 0.638 & 0.874 & 0.884 & 0.891 & 0.931 & 0.932 & $\bm{0.933}$

 \\
& eps = 1e-2 &0.652 & 0.638 & 0.678 & 0.872 & 0.887 & 0.822 & 0.931 & 0.931 & 0.932 \\

& eps = 1e-1  & 0.627 & $\bm{0.623}$ & $\bm{0.620}$ & 0.852 & 0.868 & $\bm{0.889}$ & 0.931 & 0.931 & 0.932 \\
\hline
\multirow{4}{*}{SCSG} 
& r=1e-3, B=2000 & 11.366 & 5.188 & 3.632 & 0.670 & 0.826 & 0.878 & 0.506 & 0.859 & 0.848 \\
& r=1e-4, B=2000 & 1.204 & 1.018 & 1.527 & 0.835 & 0.869 & 0.792 & 0.926 & 0.928 & 0.919 \\
& r=1e-3, B=20000 &  3.273 & 4.589 & 4.954 & 0.881 & 0.855 & 0.842 & 0.886 & 0.869 & 0.863 \\
& r=1e-4, B=20000 & 1.012 & 1.211 & 2.159 & 0.885 & 0.825 & 0.767 & 0.928 & 0.920 & 0.882 \\ 
\hline

\end{tabular}
\end{table}

\begin{table}
 \centering\small
 \caption{test on url}
\begin{tabular}{|c|c|c|c|c|c|c|c|c|c|c|}
\hline
 \multirow{2}{*}{Algo.} &  \multirow{2}{*}{hyper-para.} & \multicolumn{3}{|c|}{loss} & \multicolumn{3}{|c|}{prec.} & \multicolumn{3}{|c|}{auc}\\ \cline{3-11}
&  & 1 & 2 & 5 & 1 & 2 & 5 & 1 & 2 & 5\\ \hline
\multicolumn{2}{|c|}{$\bm{best}$} & 0.576 & 0.575 & 0.573 & 0.987 & 0.985 & 0.986 & 0.999 & 0.999 & 0.999 \\ \hline
GSA & / & ~0.589~ & ~0.587~ & ~0.585~ & ~0.973~ & ~0.974~ & ~0.977~ & ~0.995~ & ~0.996~ & ~0.997~\\ \hline
\multirow{5}{*}{SGD}  
& r=5 & 0.577 & 0.578 & $\bm{0.573}$ & 0.983 & 0.981 & 0.988 & 0.985 & 0.985 & 0.986
 \\
& r=1 &  $\bm{0.576}$ & 0.580 & 0.579 & 0.984 & 0.98 & 0.981 & 0.988 & 0.988 & 0.989
 \\
& r=.1 & $\bm{0.576}$ & $\bm{0.575}$ & $\bm{0.573}$ & $\bm{0.985}$ & $\bm{0.986}$ & $\bm{0.990}$ & $\bm{0.999}$ & $\bm{0.999}$ & $\bm{0.999}$
 \\
& r=.01 & 0.578 & 0.577 & 0.576 & 0.984 & $\bm{0.986}$ & 0.987 & 0.998 & 0.999 & 0.999
 \\
& r=.001 & 0.585 & 0.583 & 0.581 & 0.977 & 0.98 & 0.983 & 0.997 & 0.997 & 0.998
\\
\hline
\multirow{5}{*}{Adadelta} 
& eps = 1e-8  &  0.588 & 0.585 & 0.582 & 0.96 & 0.965 & 0.97 & 0.986 & 0.989 & 0.992
\\
& eps = 1e-6  & 0.581 & 0.580 & 0.579 & 0.973 & 0.975 & 0.978 & 0.994 & 0.995 & 0.995
 \\
& eps = 1e-4 & 0.577 & 0.576 & 0.575 & 0.982 & 0.982 & 0.985 & 0.997 & 0.997 & 0.997
 \\
& eps = 1e-2 & $\bm{0.576}$ & 0.576 & 0.577 & 0.985 & $\bm{0.985}$ & 0.984 & 0.994 & 0.995 & 0.995
 \\
& eps = 1e-1  & 0.579 & $\bm{0.575}$ & 0.573 & 0.980 & 0.983 & 0.989 & 0.988 & 0.987 & 0.991 \\
\hline

\end{tabular}
\end{table}

\begin{table}
 \centering\small
 \caption{test on letter.scale}
\begin{tabular}{|c|c|c|c|c|c|c|c|}
\hline
 \multirow{2}{*}{Algo.} &  \multirow{2}{*}{hyper-para.} & \multicolumn{3}{|c|}{loss} & \multicolumn{3}{|c|}{prec.} \\ \cline{3-8}
&  & 1 & 2 & 10 & 1 & 2 & 10 \\ \hline
\multicolumn{2}{|c|}{$\bm{best}$} & 1.040 & 0.964 & 0.866 & 0.713 & 0.740 & 0.770 \\ \hline
GSA & / & ~~$\bm{1.040}$~~ & ~~$\bm{0.964}$~~ & ~~0.940~~ & ~~$\bm{0.713}$~~ & ~~0.728~~ & ~~0.735~~ \\ \hline
\multirow{4}{*}{SGD}  
& r=5 &  4.966 & 4.512 & 5.825 & 0.635 & 0.663 & 0.610 \\
& r=1 &  1.303 & 1.156 & 1.333 & 0.665 & 0.702 & 0.679 \\
& r=.1 & 1.204 & 1.052 & 0.947 & 0.684 & 0.729 & 0.742 \\
& r=.01 & 2.153 & 1.777 & 1.380 & 0.521 & 0.628 & 0.687\\
& r=.001 & 3.060 & 2.893 & 2.517 & 0.294 & 0.360 & 0.478 \\
\hline
\multirow{5}{*}{Adadelta} 
& eps = 1e-8  & 3.258 & 3.258 & 3.257 & 0.210 & 0.220 & 0.194 \\
& eps = 1e-6  & 3.231 & 3.202 & 3.114 & 0.215 & 0.230 & 0.231 \\
& eps = 1e-4 &  2.098 & 1.699 & 1.306 & 0.548 & 0.618 & 0.686\\
& eps = 1e-2 &  1.091 & 1.137 & 1.071 & 0.682 & 0.664 & 0.711\\
& eps = 1e-1  & 1.330 & 1.480 & 1.402 & 0.636 & 0.624 & 0.655 \\
\hline
\multirow{6}{*}{SCSG} 
& r=5, B=500 &  19.683 & 17.000 & 12.479 & 0.126 & 0.207 & 0.376\\
& r=1, B=500 &  3.058 & 2.511 & 1.896 & 0.115 & 0.280 & 0.529\\
& r=.1, B=500 & 3.302 & 2.632 & 1.918 & 0.066 & 0.232 & 0.525\\
& r=5, B=5000 &  3.873 & 7.203 & 6.825 & 0.311 & 0.471 & 0.584\\
& r=1, B=5000 &  1.369 & 1.029 & $\bm{0.866}$ & 0.604 & $\bm{0.740}$ & $\bm{0.770}$\\
& r=.1, B=5000 & 1.767 & 1.372 & 1.070 & 0.531 & 0.680 & 0.745\\
\hline
\end{tabular}
\end{table}

\begin{table}
 \centering\small
 \caption{test on dna.scale}
\begin{tabular}{|c|c|c|c|c|c|c|c|}
\hline
 \multirow{2}{*}{Algo.} &  \multirow{2}{*}{hyper-para.} & \multicolumn{3}{|c|}{loss} & \multicolumn{3}{|c|}{prec.} \\ \cline{3-8}
&  & 1 & 2 & 10 & 1 & 2 & 10 \\ \hline
\multicolumn{2}{|c|}{$\bm{best}$} & 0.266 & 0.211 & 0.173 & 0.921 & 0.935 & 0.943 \\ \hline
GSA & / & ~~0.292~~ & ~~0.235~~ & ~~0.198~~ & ~~$\bm{0.921}$~~ & ~~$\bm{0.935}$~~ & ~~$\bm{0.943}$~~ \\ \hline
\multirow{5}{*}{SGD}  
& r=5 &  nan & nan & nan & 0.255 & 0.255 & 0.255 \\
& r=1 &   2.068 & 1.803 & 1.442 & 0.895 & 0.916 & 0.933\\
& r=.1 &  $\bm{0.266}$ & 0.246 & 0.218 & 0.909 & 0.927 & 0.940\\
& r=.01 & 0.280 & $\bm{0.211}$ & $\bm{0.173}$ & 0.920 & $\bm{0.935}$ & 0.941\\
& r=.001 & 0.687 & 0.525 & 0.350 & 0.826 & 0.890 & 0.917\\
\hline
\multirow{5}{*}{Adadelta} 
& eps = 1e-8  &  1.098 & 1.098 & 1.096 & 0.508 & 0.508 & 0.508\\
& eps = 1e-6  & 1.063 & 1.029 & 0.956 & 0.508 & 0.508 & 0.508 \\
& eps = 1e-4 &  0.461 & 0.317 & 0.215 & 0.893 & 0.922 & 0.939\\
& eps = 1e-2 & 0.726 & 0.772 & 0.626 & 0.874 & 0.900 & 0.933 \\
& eps = 1e-1  &  2.513 & 2.588 & 1.517 & 0.843 & 0.859 & 0.926\\
\hline
\multirow{6}{*}{SCSG} 
& r=1, B=200 &  7.604 & 5.778 & 3.489 & 0.642 & 0.713 & 0.830\\
& r=.1, B=200 & 2.445 & 1.260 & 0.537 & 0.556 & 0.695 & 0.843\\
& r=.01, B=200 &4.225 & 3.672 & 2.485 & 0.379 & 0.413 & 0.531 \\
& r=1, B=1000 & 9.890 & 6.046 & 3.660 & 0.618 & 0.755 & 0.853 \\
& r=.1, B=1000 & 0.995 & 1.801 & 0.270 & 0.736 & 0.620 & 0.917\\
& r=.01, B=1000 & 2.327 & 1.562 & 0.669 & 0.564 & 0.650 & 0.807\\
\hline
\end{tabular}
\end{table}

\begin{table}
 \centering\small
 \caption{test on sector.scale}
\begin{tabular}{|c|c|c|c|c|c|c|c|}
\hline
 \multirow{2}{*}{Algo.} &  \multirow{2}{*}{hyper-para.} & \multicolumn{3}{|c|}{loss} & \multicolumn{3}{|c|}{prec.} \\ \cline{3-8}
&  & 1 & 2 & 10 & 1 & 2 & 10 \\ \hline
\multicolumn{2}{|c|}{$\bm{best}$} & 1.070 & 0.727 & 0.545 & 0.806 & 0.890 & 0.920 \\ \hline
GSA & / & ~~1.356~~ & ~~0.950~~ & ~~0.728~~ & ~~0.794~~ & ~~$\bm{0.890}$~~ & ~~$\bm{0.920}$~~ \\ \hline
\multirow{4}{*}{SGD}  
& r=20 & 1.441 & 1.985 & 1.003 & 0.803 & 0.756 & 0.882 \\
& r=5 &   $\bm{1.070}$ & $\bm{0.727}$ & $\bm{0.545}$ & $\bm{0.806}$ & 0.872 & 0.916 \\
& r=1 &  2.684 & 1.662 & 0.911 & 0.558 & 0.829 & 0.910 \\
& r=.1 &  4.422 & 4.161 & 3.385 & 0.180 & 0.140 & 0.687\\
\hline
\multirow{5}{*}{Adadelta} 
& eps = 1e-8  & 4.654 & 4.654 & 4.653 & 0.358 & 0.502 & 0.552 \\
& eps = 1e-6  & 4.645 & 4.634 & 4.596 & 0.360 & 0.504 & 0.562 \\
& eps = 1e-4 & 4.383 & 4.058 & 3.106 & 0.370 & 0.534 & 0.761 \\
& eps = 1e-2 & 2.882 & 1.808 & 0.961 & 0.681 & 0.845 & 0.915 \\
& eps = 1e-1  & 2.674 & 1.678 & 0.921 & 0.677 & 0.838 & 0.914 \\
\hline
\multirow{6}{*}{SCSG} 
& r=5, B=500 & 4.171 & 3.363 & 2.015 & 0.131 & 0.327 & 0.683 \\
& r=1, B=500 &4.969 & 4.697 & 4.072 & 0.028 & 0.045 & 0.148 \\
& r=.1, B=500 & 5.325 & 5.199 & 5.057 & 0.010 & 0.012 & 0.012\\
& r=5, B=3000 & 2.203 & 1.935 & 0.861 & 0.592 & 0.650 & 0.857 \\
& r=1, B=3000 & 4.000 & 3.059 & 1.813 & 0.167 & 0.419 & 0.713 \\
& r=.1, B=3000 & 5.049 & 4.882 & 4.512 & 0.012 & 0.018 & 0.061\\
\hline
\end{tabular}
\end{table}

\begin{table}
 \centering\small
 \caption{test on usps}
\begin{tabular}{|c|c|c|c|c|c|c|c|}
\hline
 \multirow{2}{*}{Algo.} &  \multirow{2}{*}{hyper-para.} & \multicolumn{3}{|c|}{loss} & \multicolumn{3}{|c|}{prec.} \\ \cline{3-8}
&  & 1 & 2 & 5 & 1 & 2 & 5 \\ \hline
\multicolumn{2}{|c|}{$\bm{best}$} & 0.367  & 0.369 & 0.354 & 0.902 & 0.911 & 0.912 \\ \hline
GSA & / & ~~0.392~~ & ~~0.378~~ & ~~$\bm{0.354}$~~ & ~~0.898~~ & ~~0.897~~ & ~~0.909~~ \\ \hline
\multirow{4}{*}{SGD}  
& r=1 &  nan & nan & nan & 0.179 & 0.179 & 0.179 \\
& r=.1 &  1.348 & 1.325 & 1.597 & 0.891 & 0.898 & 0.894\\
& r=.01 & $\bm{0.367}$ & 0.374 & 0.376 & $\bm{0.902}$ & $\bm{0.911}$ & $\bm{0.912}$\\
& r=.001 & 0.492 & 0.423 & 0.365 & 0.879 & 0.890 & 0.906\\
\hline
\multirow{5}{*}{Adadelta} 
& eps = 1e-8  &  2.293 & 2.281 & 2.245 & 0.322 & 0.331 & 0.332\\
& eps = 1e-6  & 1.650 & 1.270 & 0.807 & 0.608 & 0.740 & 0.835 \\
& eps = 1e-4 & 0.377 & $\bm{0.369}$ & 0.357 & 0.898 & 0.901 & 0.908 \\
& eps = 1e-2 & 2.880 & 3.263 & 2.686 & 0.877 & 0.870 & 0.900 \\
& eps = 1e-1  & nan & nan & nan & 0.179 & 0.179 & 0.179 \\
\hline
\multirow{5}{*}{SCSG} 
& r=.1, B=500 & 10.544 & 4.763 & 3.606 & 0.299 & 0.649 & 0.739\\
& r=.01, B=500 & 4.214 & 2.634 & 1.327 & 0.380 & 0.543 & 0.736\\
& r=.1, B=3000 & 2.257 & 2.151 & 1.464 & 0.814 & 0.837 & 0.893 \\
& r=.01, B=3000 & 1.992 & 0.921 & 0.599 & 0.508 & 0.796 & 0.875\\
& r=.001, B=3000 &  1.758 & 1.235 & 0.624 & 0.610 & 0.704 & 0.863\\
\hline
\end{tabular}
\end{table}

\begin{table}
 \centering\small
 \caption{test on protein}
\begin{tabular}{|c|c|c|c|c|c|c|c|}
\hline
 \multirow{2}{*}{Algo.} &  \multirow{2}{*}{hyper-para.} & \multicolumn{3}{|c|}{loss} & \multicolumn{3}{|c|}{prec.} \\ \cline{3-8}
&  & 1 & 2 & 5 & 1 & 2 & 5 \\ \hline
\multicolumn{2}{|c|}{$\bm{best}$} & 0.777 & 0.763 & 0.762 & 0.676 & 0.682 & 0.687 \\ \hline
GSA & / & ~~0.805~~ & ~~0.810~~ & ~~0.794~~ & ~~0.667~~ & ~~0.669~~ & ~~0.679~~ \\ \hline
\multirow{5}{*}{SGD}  
& r=5 &  10.403 & 10.968 & 10.476 & 0.614 & 0.592 & 0.614\\
& r=1 &   4.042 & 4.315 & 4.116 & 0.597 & 0.597 & 0.611\\
& r=.1 &  0.895 & 0.888 & 0.886 & 0.644 & 0.651 & 0.659\\
& r=.01 & $\bm{0.777}$ & $\bm{0.763}$ & $\bm{0.762}$ & $\bm{0.676}$ & $\bm{0.682}$ & $\bm{0.687}$\\
& r=.001 & 0.928 & 0.869 & 0.796 & 0.602 & 0.648 & 0.681\\
\hline
\multirow{5}{*}{Adadelta} 
& eps = 1e-8  & 1.098 & 1.096 & 1.093 & 0.470 & 0.470 & 0.470 \\
& eps = 1e-6 &  1.041 & 1.014 & 0.965 & 0.470 & 0.470 & 0.512 \\
& eps = 1e-4 &  0.781 & 0.764 & 0.770 & 0.673 & 0.680 & 0.683 \\
& eps = 1e-2 &  2.191 & 2.488 & 2.510 & 0.602 & 0.596 & 0.606 \\
& eps = 1e-1  &  3.935 & 4.048 & 3.939 & 0.588 & 0.613 & 0.620 \\
\hline
\multirow{6}{*}{SCSG} 
& r=1, B=1000 & 4.110 & 2.910 & 2.608 & 0.444 & 0.572 & 0.549 \\
& r=.1, B=1000 & 1.677 & 1.188 & 0.861 & 0.419 & 0.529 & 0.639\\
& r=.01, B=1000 &2.443 & 2.272 & 1.887 & 0.376 & 0.385 & 0.426 \\
& r=1, B=5000 &  4.644 & 3.387 & 3.708 & 0.539 & 0.594 & 0.581\\
& r=.1, B=5000 & 0.882 & 0.825 & 0.771 & 0.632 & 0.651 & 0.681\\
& r=.01, B=5000 & 2.109 & 1.726 & 0.952 & 0.482 & 0.496 & 0.611\\
\hline
\end{tabular}
\end{table}

\begin{table}
 \centering\small
 \caption{test on rcv1.multiclass}
\begin{tabular}{|c|c|c|c|c|c|c|c|}
\hline
 \multirow{2}{*}{Algo.} &  \multirow{2}{*}{hyper-para.} & \multicolumn{3}{|c|}{loss} & \multicolumn{3}{|c|}{prec.} \\ \cline{3-8}
&  & 1 & 2 & 5 & 1 & 2 & 5 \\ \hline
\multicolumn{2}{|c|}{$\bm{best}$} &0.586  & 0.530 & 0.432 & 0.853 & 0.869 & 0.886 \\ \hline
GSA & / & ~~0.659~~ & ~~0.579~~ & ~~0.492~~ & ~~0.851~~ & ~~0.865~~ & ~~0.880~~ \\ \hline
\multirow{4}{*}{SGD}  
& r=5 &  $\bm{0.586}$ & 0.591 & 0.563 & $\bm{0.853}$ & 0.850& 0.865\\
& r=1 & 0.646 & 0.539 & 0.444 & 0.851 & 0.867 & 0.881\\
& r=.1 & 1.499 & 1.157 & 0.800 & 0.705  & 0.783 & 0.847\\
& r=.01 & 2.562  & 2.311 & 1.868 &  0.326 & 0.432 & 0.622\\
\hline
\multirow{5}{*}{Adadelta} 
& eps = 1e-8  & 3.968 & 3.966 & 3.961 & 0.274 & 0.280 & 0.281\\
& eps = 1e-6  & 3.822 & 3.673 & 3.279 & 0.277 & 0.287 & 0.306 \\
& eps = 1e-4  & 1.561 & 1.116 & 0.724 & 0.673 & 0.774 & 0.845 \\
& eps = 1e-2 &  0.652 &  $\bm{0.530}$ &  $\bm{0.432}$ & 0.857 & $\bm{0.869}$ & $\bm{0.886}$\\
& eps = 1e-1 &  0.662 & 0.547 & 0.433 & 0.840 & 0.865 & 0.885 \\
\hline
\multirow{4}{*}{SCSG} 
& r=1, B=500 & 2.806 & 2.221 & 1.576 & 0.343 & 0.486 & 0.639\\
& r=.1, B=500 & 3.609 & 3.261 & 2.854 & 0.171&0.225 &0.289\\
& r=1, B=5000 &  1.351 & 0.954 & 0.667 & 0.716& 0.789&0.838\\
& r=.1, B=5000 & 3.011 & 2.426 & 1.550 & 0.286& 0.455&0.642\\
\hline
\end{tabular}
\end{table}

\end{document}